%% file: pami.tex
\documentclass[10pt,journal,compsoc]{IEEEtran}
\ifCLASSOPTIONcompsoc
  \usepackage[nocompress]{cite}
\else
  \usepackage{cite}
\fi
\ifCLASSINFOpdf
\else
\fi

\usepackage{epsfig}
\usepackage{graphicx}
\usepackage{amsmath}
\usepackage{amssymb}
\usepackage{enumitem}
\usepackage{ragged2e}

\usepackage[pagebackref=true,breaklinks=true,letterpaper=true,colorlinks,bookmarks=false,urlcolor=red,citecolor=cyan]{hyperref}
\usepackage{gensymb,graphics,subfigure,inputenc}
\usepackage[linesnumbered,algo2e,boxed]{algorithm2e}
\graphicspath{{Figure/}}
\usepackage[figuresright]{rotating}
\usepackage{amsmath,amssymb,textcomp,subfigure,multirow,upgreek}
\usepackage{booktabs}
\usepackage{color,mdwlist}

\usepackage{graphicx,fontawesome}
\graphicspath{{Figures/}}

\usepackage{caption}
\usepackage{enumitem}
\setitemize{noitemsep,topsep=0pt,parsep=0pt,partopsep=0pt}

\usepackage{arydshln}

\begin{document}
%
\title{Enhanced Multi-Scale Cross-Attention for Person Image Generation}
\author{Hao Tang,
	 Ling Shao,
	Nicu Sebe,
        Luc Van Gool
	\IEEEcompsocitemizethanks{
	   \IEEEcompsocthanksitem 
        Hao Tang is with the School of Computer Science, Peking University, Beijing 100871, China. E-mail: haotang@pku.edu.cn \protect
        \IEEEcompsocthanksitem 
        Ling Shao is with the UCAS-Terminus AI Lab, University of Chinese Academy of Sciences, Beijing 100049, China \protect
	\IEEEcompsocthanksitem Nicu Sebe is with the Department of Information Engineering and Computer Science (DISI), University of Trento, Trento 38123, Italy.  \protect
        \IEEEcompsocthanksitem Luc Van Gool is with the Department of Information Technology and
        Electrical Engineering, ETH Zurich, Switzerland, with the Department of Electrical Engineering, KU Leuven, Belgium, and with INSAIT, Sofia Un., Bulgaria.\protect
        }
	\thanks{Corresponding authors: Hao Tang, Ling Shao.}}

%
%

\markboth{IEEE Transactions on Pattern Analysis and Machine Intelligence}%
{Shell \MakeLowercase{\textit{et al.}}: Bare Demo of IEEEtran.cls for Computer Society Journals}
%



\IEEEtitleabstractindextext{%
\input{0abstract}

}

\maketitle

\IEEEdisplaynontitleabstractindextext

%
\IEEEpeerreviewmaketitle


%
%
%
%

\input{1introduction}
\input{2relatedworks}
\input{3method}
\input{4experiments}
\input{5conclusions}

\bibliographystyle{IEEEtran}
\bibliography{reference}

\begin{IEEEbiography}[{\includegraphics[width=1in,height=1.25in,clip,keepaspectratio]{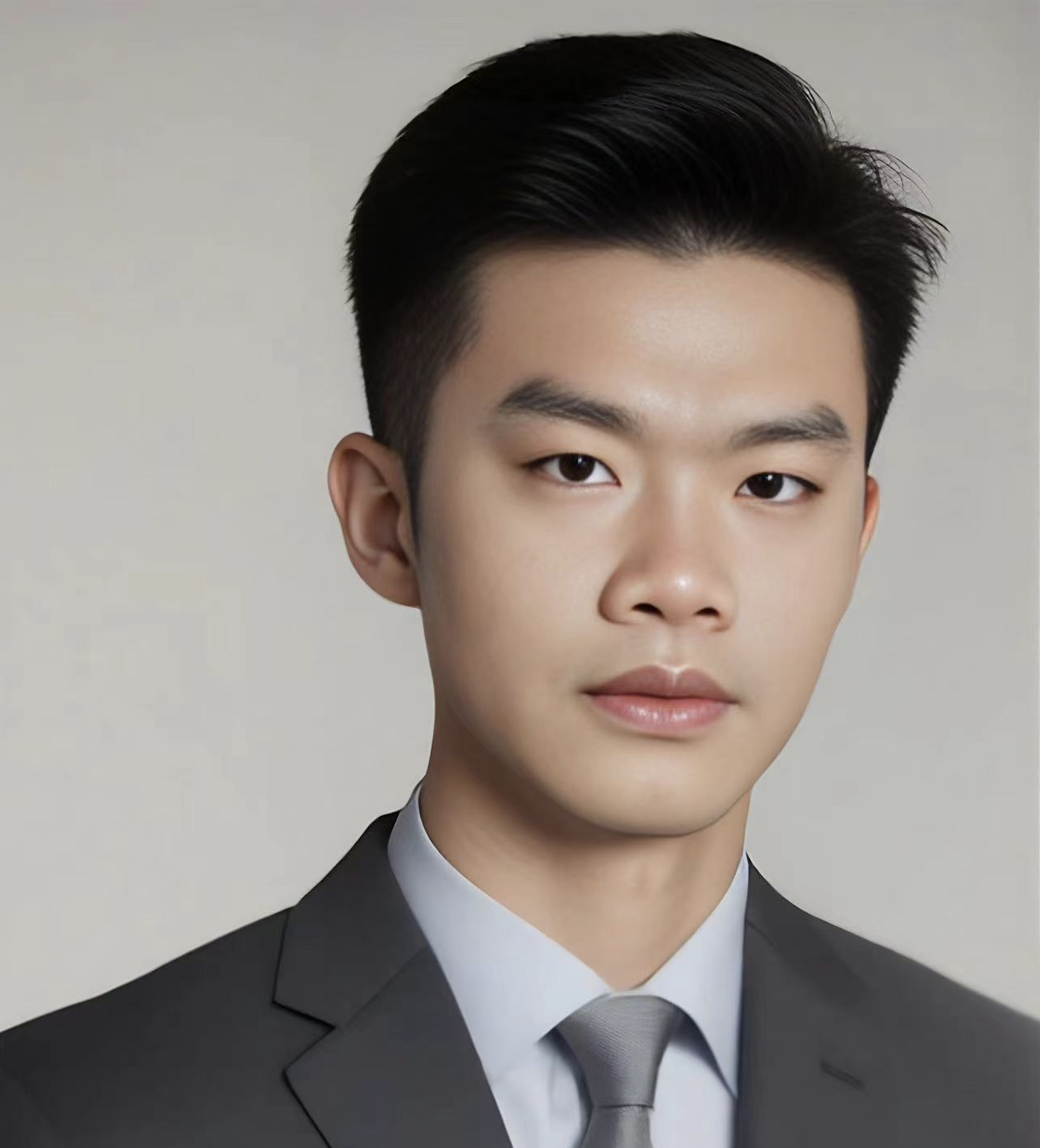}}]{Hao Tang}
is a tenure-track Assistant Professor at Peking University, China. Previously, he held postdoctoral positions at both CMU, USA, and ETH Zürich, Switzerland. He earned a master's degree from Peking University, China, and a Ph.D. from the University of Trento, Italy. He was a visiting Ph.D. student at the University of Oxford, UK, and an intern at IIAI, UAE. His research interests include AIGC, LLM, machine learning, computer vision, embodied AI, and their applications to scientific domains.
\end{IEEEbiography}

\begin{IEEEbiography}[{\includegraphics[width=1in,height=1.25in,clip,keepaspectratio]{
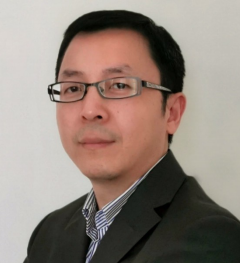}}]{Ling Shao} is a Distinguished Professor with the UCAS-Terminus AI Lab, University of Chinese Academy of Sciences, Beijing, China. He was the founder of the Inception Institute of Artificial Intelligence (IIAI) and the Mohamed bin Zayed University of Artificial Intelligence (MBZUAI), Abu Dhabi, UAE. His research interests include generative AI, vision and language, and AI for healthcare. He is a fellow of the IEEE, the IAPR, the BCS and the IET.
\end{IEEEbiography}

\begin{IEEEbiography}[{\includegraphics[width=1in,height=1.25in,clip,keepaspectratio]{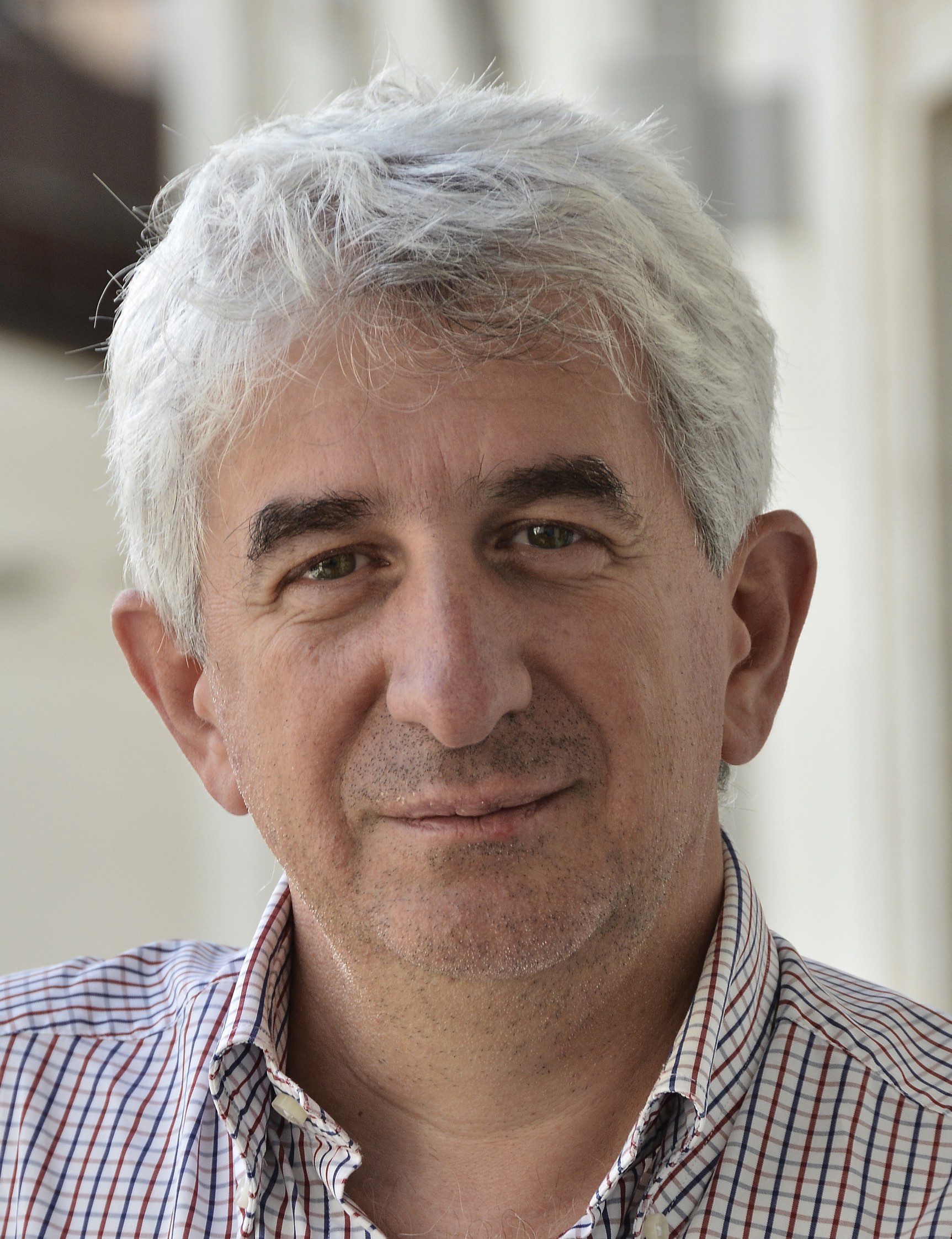}}]{Nicu Sebe} is Professor with the University of
Trento, Italy, leading the research in the areas of multimedia information retrieval and human behavior understanding. He was the General Chair of ACM Multimedia 2013, and the Program Chair of ACM Multimedia 2007 and 2011, ECCV 2016, ICCV 2017 and ICPR 2020. He is a fellow of the International Association for
Pattern Recognition and of ELLIS.
\end{IEEEbiography}

\begin{IEEEbiography}[{\includegraphics[width=1in,height=1.25in,clip,keepaspectratio]{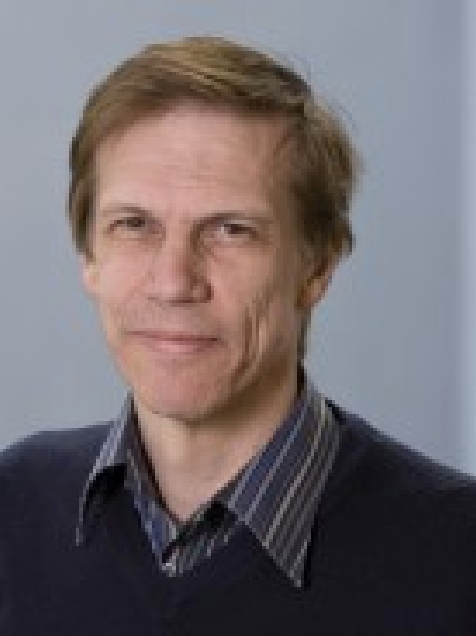}}]{Luc Van Gool} is full professor for computer vision at INSAIT, Sofia Un.
He is professor emeritus at ETH Zurich and the KU Leuven. He has been a
program committee member of several, major computer vision conferences
(e.g. Program Chair ICCV'05, Beijing, General Chair of ICCV'11,
Barcelona, and of ECCV'14, Zurich). His main interests include 3D
reconstruction and modeling, object recognition, and autonomous driving.
He is a co-founder of more than 10
spin-off companies. 
\end{IEEEbiography}




\end{document}

%% file: 0abstract.tex
\justify

\begin{abstract}
In this paper, we propose a novel cross-attention-based generative adversarial network (GAN) for the challenging person image generation task. Cross-attention is a novel and intuitive multi-modal fusion method in which an attention/correlation matrix is calculated between two feature maps of different modalities. Specifically, we propose the novel XingGAN (or CrossingGAN), which consists of two generation branches that capture the person's appearance and shape, respectively. Moreover, we propose two novel cross-attention blocks to effectively transfer and update the person's shape and appearance embeddings for mutual improvement. This has not been considered by any other existing GAN-based image generation work.  To further learn the long-range correlations between different person poses at different scales and sub-regions, we propose two novel multi-scale cross-attention blocks. To tackle the issue of independent correlation computations within the cross-attention mechanism leading to noisy and ambiguous attention weights, which hinder performance improvements, we propose a module called enhanced attention (EA). Lastly, we introduce a novel densely connected co-attention module to fuse appearance and shape features at different stages effectively. 
Extensive experiments on two public datasets demonstrate that the proposed method outperforms current GAN-based methods and performs on par with diffusion-based methods. However, our method is significantly faster than diffusion-based methods in both training and inference.

\end{abstract}

%% file: 1introduction.tex
\section{Introduction}

\begin{figure*}[t]
	\centering
	\includegraphics[width=1\linewidth]{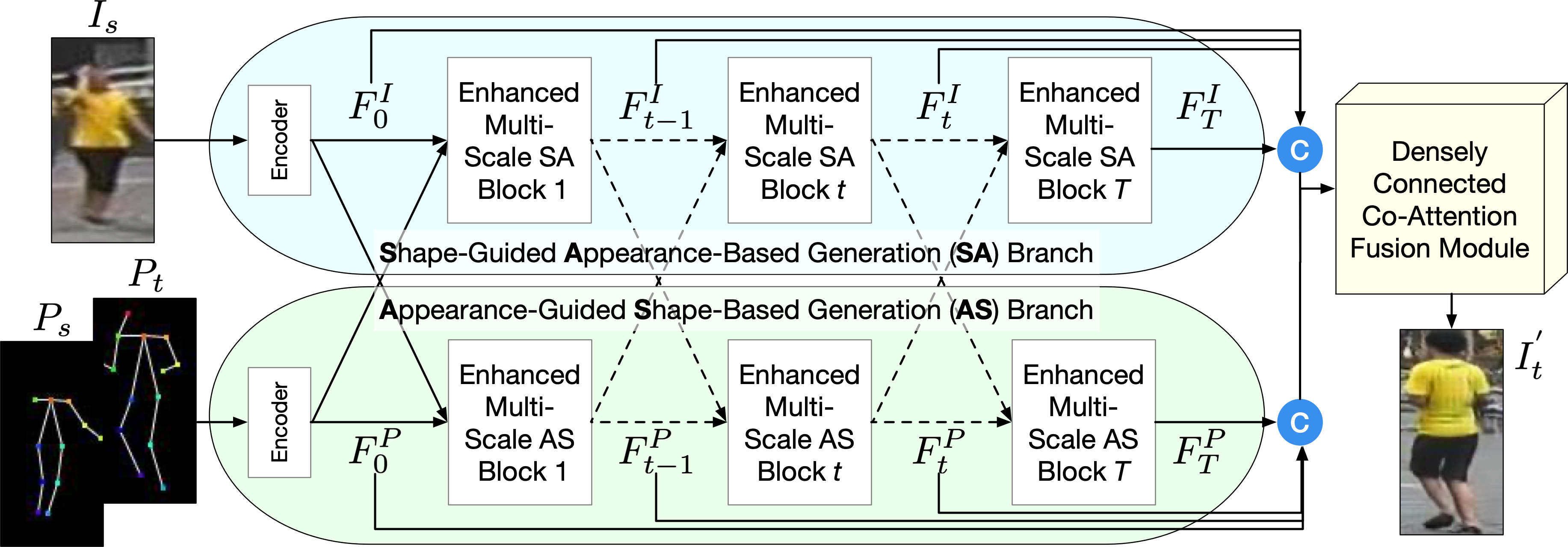}
	\caption{Overview of the proposed XingGAN++ for the person image generation task. Both the shape-guided appearance-based generation (SA) and the appearance-guided shape-based generation (AS) branches consist of a sequence of interweaved enhanced multi-scale SA and AS blocks.
	For brevity, we omit both the appearance-guided and shape-guided discriminator in the figure. All these components are trained in an end-to-end fashion so that the SA and AS branches can benefit from each other to generate more shape-consistent and appearance-consistent person images.}
	\label{fig:method}
	\vspace{-0.4cm}
\end{figure*}

\IEEEPARstart{T}{he} goal of person image generation is to generate photorealistic person images conditioned on an input person image and several desired poses.
This task has a wide range of applications, such as person image/video generation \cite{yang2018pose,grigorev2019coordinate,balakrishnan2018synthesizing,ilyes2018pose,liu2019liquid} and person re-identification \cite{zhu2019progressive,qian2018pose,guo2022unsupervised,wu2022identity}.
Existing methods such as \cite{ma2017pose,ma2018disentangled,siarohin2018deformable,zhu2019progressive,tang2019cycle,tang2022bipartite} have achieved promising performance on person image generation.
For example, Zhu et al.~\cite{zhu2019progressive} recently proposed a conditional generative adversarial network (GAN) that comprises a sequence of pose-attentional transfer blocks for person image generation. Each block transfers certain regions it attends to and progressively generates the desired person image.
Moreover, Zhang et al. \cite{zhang2022exploring} proposed a dual-task pose transformer network (DPTN), which introduces an auxiliary task (i.e., source-to-source task) to promote the generation performance.

Although \cite{zhu2019progressive,han2018viton,wang2018toward,minar2020cp,zhang2022exploring,roy2023multi,wang2023novel} have advanced our understanding of the person image generation task, we still observe unsatisfactory results and visual artifacts in their generated images for several reasons.
First, \cite{zhu2019progressive} stacks several convolutional layers to generate the attention maps of the shape features, which are then used to attentively highlight the appearance features.
Since convolutional operations are building blocks that process one local neighborhood at a time, this means that they cannot capture the joint influence of the appearance and the shape features.
Second, the attention maps in \cite{zhu2019progressive} are produced using a single modality, i.e., the pose, leading to insufficiently accurate correlations for both modalities (i.e., the pose and the image), and thus misguiding the image generation. Third, existing such as \cite{zhu2019progressive,han2018viton,wang2018toward,minar2020cp,zhang2022exploring} do not consider the long-range global interactive correlations between person poses at different scales and sub-regions, leading to unsatisfactory and inconsistent results.

Based on these observations, we propose two novel cross-attention-based networks, i.e., XingGAN and XingGAN++, for person image generation.
Note that cross-attention is a novel and intuitive multi-modal fusion method in which an attention/correlation matrix is calculated between two feature maps of different modalities.
Specifically, we propose the XingGAN framework, which consists of an Xing generator, a shape-guided discriminator, and an appearance-guided discriminator.
The Xing generator consists of three components, i.e., a shape-guided appearance-based generation (SA) branch, an appearance-guided shape-based generation (AS) branch, and a co-attention fusion module.
The proposed SA branch contains a sequence of SA blocks aiming to progressively update the appearance representation under the guidance of the shape representation. In contrast, the proposed AS branch contains a sequence of AS blocks aiming to progressively update the shape representation under the guidance of the appearance representation. 
We also present a novel cross-operation in both the SA and AS blocks to capture the joint influence of the image modality and the pose modality by creating attention maps produced by both modalities jointly. 
Moreover, we introduce a co-attention fusion module to better fuse the final appearance and shape features and generate the desired person images.
Finally, we present an appearance-guided and a shape-guided discriminator to simultaneously judge how likely the generated image is to contain the same person as the input image and how well the generated image aligns with the target pose, respectively.
The proposed XingGAN is trained in an end-to-end fashion so that the generation branches can enjoy mutual benefits from each other.

To further improve the performance, we propose XingGAN++, which contains three new contributions based on XingGAN.
The overall framework of the proposed XingGAN++ is shown in Figure~\ref{fig:method}.
(i) Since there exists a large deformation between the source and the target poses, a single-scale feature may not be able to capture all the necessary spatial information for a fine-grained generation. 
Therefore, we propose two novel multi-scale cross-attention blocks to capture the correlations between person poses at different scales and sub-regions.
In this way, we obtain multi-scale cross-attention with different receptive fields to perceive different spatial contexts, leading to better generation performance.
(ii) The correlation for each query-key pair is calculated independently, ignoring the correlations of other query-key pairs. This can introduce false correlations due to imperfect feature representations or the presence of distracting image patches in cluttered background scenes, leading to noisy and ambiguous attention weights. 
To tackle this problem, we propose a new enhanced attention (EA) module, which builds upon the traditional attention mechanism by incorporating an inner attention module. The purpose of this inner attention module is to enhance correlation by seeking consensus among all associated vectors. Put simply, when a query exhibits a strong correlation with a key, it is likely that some neighboring keys will also display a relatively strong correlation. Conversely, if there is no consensus among the keys, the correlation might be considered as noise. Drawing inspiration from this idea, we introduce an enhanced attention module to leverage these message notifications.
(iii) Lastly, we introduce a novel densely connected co-attention fusion module to effectively fuse appearance and shape features at different stages, leading to more photorealistic and consistent results. The proposed XingGAN++ is also trained in an end-to-end manner.

To summarize, the contributions of our paper are:
\begin{itemize}[leftmargin=*]
	\item We propose two novel methods (i.e., XingGAN and XingGAN++) for the person image generation task. Both explore cascaded guidance with two different generation branches and aim to progressively produce a more detailed image from both person shape and appearance embeddings.
	\item We propose SA and AS blocks, which effectively transfer and update person shape and appearance modalities through a cross-operation for mutual improvement and can significantly boost the quality of the final person images.
        \item We further propose two novel multi-scale SA and AS blocks, which aim to capture the long-range correlations between person poses at different scales and sub-regions.
        \item We propose a novel enhanced attention (EA) module that addresses the shortcomings of traditional attention mechanisms by reducing noise and ambiguity in correlation maps. By incorporating this module, we can achieve substantial improvements in generation performance.
        \item We introduce a novel densely connected co-attention fusion module to effectively fuse appearance and shape features at different stages.
	\item Experiments on two public datasets (i.e., Market-1501 \cite{zheng2015scalable} and DeepFashion \cite{liu2016deepfashion}) show that the proposed method surpasses existing GAN-based methods and achieves comparable performance to diffusion-based methods. Furthermore, our method is considerably faster than diffusion-based methods in both training and inference.
\end{itemize}

Part of the material presented here appeared in our conference version \cite{tang2020xinggan}. The current paper extends \cite{tang2020xinggan} in several ways.
(1) We present a more detailed analysis of related works by including recently published works on the person image generation task.
(2) To solve the different-scale problem of the same person in different scenes, we propose two novel multi-scale cross-attention blocks to capture the correlations between person poses at different scales and sub-regions.
(3) We propose a novel enhanced attention (EA) module designed to strengthen accurate correlations while suppressing false ones by fostering consensus among all correlation vectors.
(4) We extend the co-attention fusion module proposed in \cite{tang2020xinggan} to a more robust and effective densely connected co-attention fusion module, which aims to fuse the appearance and shape features at different stages. 
Equipped with these three newly proposed components, our XingGAN proposed in \cite{tang2020xinggan} is upgraded to XingGAN++.
(5) We conduct extensive ablation studies to demonstrate the effectiveness of the proposed methods.
(6) We extend the experiments by comparing our method with very recent GAN-based and diffusion model-based works on two public datasets. We observe that the proposed method achieves much better results than existing GAN-based methods. At the same time, our results are on par with those of diffusion-based methods (i.e., PIDM \cite{bhunia2023person}), but our approach is significantly faster in both training and inference. Specifically, the proposed method is 69.33$\times$ faster than PIDM during training and 18.04$\times$ faster during inference.

%% file: 2relatedworks.tex
\section{Related Work}
\noindent \textbf{Generative Adversarial Networks (GANs)} \cite{goodfellow2014generative} consist of a generator and a discriminator, where the goal of the generator is to produce photorealistic images that the discriminator cannot tell apart from real images.
GANs have demonstrated remarkable ability in generating realistic images \cite{brock2018large,karras2018style,shaham2019singan,tao2022net,zhang2022unsupervised,zhang20223d,tao2022df,tang2021attentiongan,tang2019expression,tang2019attention}.
However, vanilla GANs still struggle to generate images in a controlled setting.
To address this, conditional GANs (CGANs) \cite{mirza2014conditional} have been proposed.

\noindent \textbf{Image-to-Image Translation} aims to learn a translation mapping between target and input images. CGANs have achieved decent results in pixel-wise aligned image-to-image translation tasks \cite{isola2017image,tang2018gesturegan,albahar2019guided,zhang2021pise}.
For example, Isola et al. proposed Pix2pix~\cite{isola2017image}, which adopts CGANs to generate the target domain images based on the input domain images, such as photo-to-map, sketch-to-image, and night-to-day. 
However, pixel-wise alignment is not suitable for either person image generation or virtual try-on due to the shape deformation between the input person image and target person image.

\noindent \textbf{Person Image Generation.} To remedy this, several works have started to use poses to guide person image generation \cite{ma2017pose,ma2018disentangled,siarohin2018deformable,esser2018variational,tang2019cycle,zhu2019progressive,liu2021liquid,ren2022neural,cheong2022kpe}.
For example, Ma et al. introduced PG2 \cite{ma2017pose}, which is a two-stage model that generates the target person images based on an input image and the target poses.
Moreover, Ma et al. also proposed DPIG \cite{ma2018disentangled} based on PG2 \cite{ma2017pose}, which considers three factors when generating novel person images, i.e., foreground, background, and pose of the input image. Though this model provides more controllability, the quality of the generated images significantly degrades.
Siarohin et al. proposed PoseGAN \cite{siarohin2018deformable}, which requires an extensive affine transformation computation to deal with the input-output misalignment caused by pose differences.
Essner et al. proposed VUNet \cite{esser2018variational}, which tries to disentangle appearance and pose. However, several appearance misalignments are observed in the generated images.
Tang et al. introduced C2GAN \cite{tang2019cycle}, which is a cross-modal framework that explores the joint exploitation of human keypoints and images in an interactive manner. 
Zhu~et al. proposed PATN \cite{zhu2019progressive}, which contains a sequence of pose-attentional transfer blocks for generating the target person image progressively. 
Zhang et al. \cite{zhang2023adding} introduced ControlNet, an innovative neural network architecture designed to integrate spatial conditioning controls into large, pretrained text-to-image diffusion models. This approach enhances the flexibility and precision of diffusion models by allowing them to incorporate spatial information, thereby improving their ability to generate more accurate and contextually relevant images based on textual descriptions.
Bhunia et al. \cite{bhunia2023person} introduced the person image diffusion model (PIDM), which aims to decompose the intricate transfer problem into a sequence of more straightforward forward-backward denoising steps. Although PIDM has achieved impressive results, its training and inference processes are notably slow. This significant drawback limits the practicality and application scenarios of diffusion model-based methods, particularly in environments where computational resources are constrained or real-time performance is critical.
While our GAN-based method achieves comparable results to PIDM, it is 69.33$\times$ faster during training and 18.04$\times$ faster during inference.

Note that the aforementioned methods adopt human keypoints or skeletons as pose guidance, which are usually extracted using OpenPose \cite{cao2017realtime}.
In addition, several works adopt DensePose \cite{neverova2018dense}, 3D pose \cite{li2019dense}, and segmented pose \cite{dong2018soft} to generate the images because they contain more information related to body depth and part segmentation, producing better results with more textural details. 
For example, Neverova et al. \cite{neverova2018dense} proposed a framework that uses DensePose \cite{alp2018densepose} as the pose representation.
Li et al. \cite{li2019dense} also proposed a method to generate person images based on dense and intrinsic 3D appearance flow.
Dong et al. \cite{dong2018soft} first generate the target parsing using a pose-guided parser and then generate the final person images.
However, the keypoint-based pose representation is much cheaper and more flexible than the DensePose, 3D pose, and segmented pose representations and can be more easily used in practical applications.
Therefore, we favor keypoint-based pose representation in this paper.

\noindent \textbf{Image-Guidance Conditioning Schemes.}
Recently, many schemes have been proposed to incorporate extra guidance (e.g., human poses \cite{ma2017pose,zhu2019progressive,tang2020unified,tang2020bipartite}, segmentation maps \cite{park2019semantic,tang2019multi,ren2021cascaded,tang2020local,wu2022cross,wu2022cross_tmm,tang2022local,tang2021layout,liu2021cross,tang2020dual,liu2020exocentric}, facial landmarks \cite{tang2019cycle,zakharov2019few,tang2022facial,tang2021total}, etc.) into image-to-image translation models.
These can be divided into four categories, 
i.e., input concatenation \cite{tang2019cycle,xian2018texturegan,zhang2017real}, feature concatenation \cite{ma2017pose,ma2018disentangled,esser2018variational,li2019dense,lai2018learning,li2019joint}, one-way guidance-to-image interaction \cite{siarohin2018deformable,park2019semantic,huang2017arbitrary,perez2018film}, and two-way guidance-and-image interaction~\cite{zhu2019progressive,albahar2019guided,chi2019two}.

\begin{figure*}[t] \small
	\centering
	\includegraphics[width=0.8\linewidth]{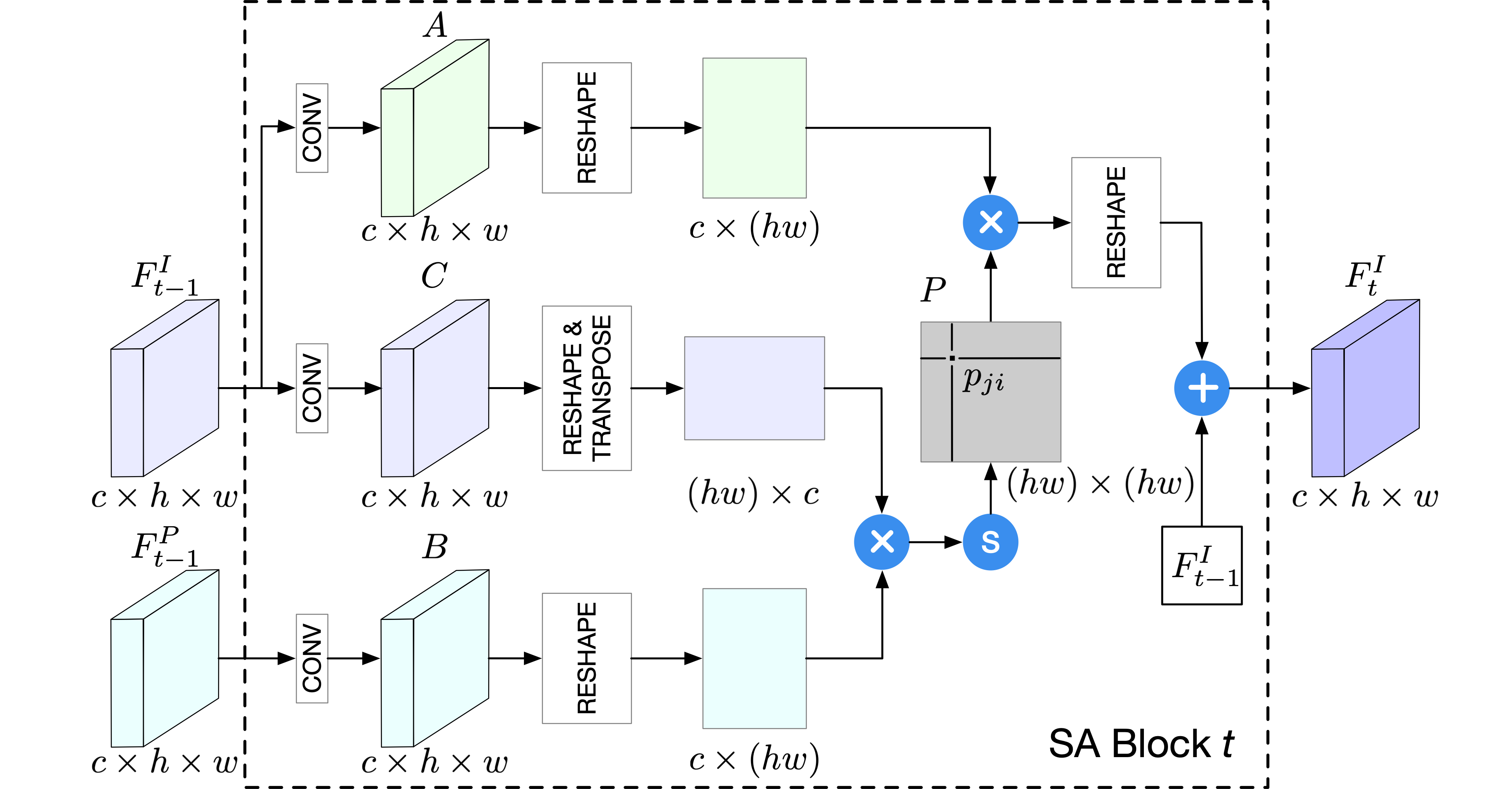}
	\caption{Structure of the proposed shape-guided appearance-based generation (SA) block (proposed in our conference version \cite{tang2020xinggan}), which takes the previous appearance code $F_{t-1}^I$ and the previous shape code $F_{t-1}^P$ as input and obtains the appearance code $F_{t}^I$ in a crossed non-local way. The symbols $\oplus$, $\otimes$, and $\textcircled{s}$ denote element-wise addition, element-wise multiplication, and Softmax activation, respectively.}
	\label{fig:sa}
	\vspace{-0.4cm}
\end{figure*}

The most straightforward way of conditioning the guidance is to concatenate it with the input image along the channel dimension.
For example, C2GAN \cite{tang2019cycle} takes the input person image and the target poses as input and outputs the corresponding target person images.
Instead of concatenating the guidance and the image in the input, other works \cite{ma2017pose,ma2018disentangled,esser2018variational} concatenate their feature representations in a certain layer.
For instance, PG2 \cite{ma2017pose} concatenates the embedded pose feature with the embedded image feature in the fully connected bottleneck layer. 
Another more general scheme is to use the guidance to guide the generation of the image.
For example, 
Park et al~\cite{park2019semantic} proposed using the input semantic labels for modulating the activations in normalization layers to generate better images.
Siarohin et al.~\cite{siarohin2018deformable} first learn an affine transformation between the input and the target pose and then use it to `move' the feature maps between the input and target image.
Unlike existing one-way guidance-to-image interaction schemes, which allow information to flow only from the guidance to the input image, a recent scheme, i.e., two-way guidance-and-image interaction, also considers the information flow from the input image back to the guidance \cite{zhu2019progressive,albahar2019guided}.
For example, Zhu et al. \cite{zhu2019progressive} proposed an attention-based GAN model to simultaneously update the person's appearance and shape features under the guidance of the other and showed that the proposed two-way guidance-and-image interaction strategy leads to better performance on person image generation tasks.

Different from the existing two-way guidance-and-image interaction schemes \cite{zhu2019progressive,albahar2019guided} that allow both the image and guidance to guide and update each other in a local way, we show that the proposed cross-conditioning strategy can further improve the performance of person image generation task.

%% file: 3method.tex
\section{The Proposed Methods}
We start by presenting the details of the proposed framework, which consists of three components, i.e., a shape-guided appearance-based generation (SA) branch modeling the person shape representation, an appearance-guided shape-based generation (AS) branch modeling the person appearance representation, and a densely connected co-attention fusion (DCCAF) module that fuses these two branches. 
In the following, we first present the design of the two proposed generation branches and then introduce the DCCAF module.
Lastly, we present our two discriminators, the overall optimization objective, and the implementation details.

The inputs of the proposed Xing generator are the source image $I_s$, the source pose $P_s$, and the target pose $P_t$.
The goal is to translate the pose of the person in the source image $I_s$ from the source pose $P_s$ to the target pose $P_t$, thus synthesizing a photorealistic person image $I_t^{'}$.
In this way, the source image $I_s$ provides the appearance information, and the poses ($P_s$, $P_t$) provide the shape information to the Xing generator for synthesizing the desired person image. 

\noindent \textbf{Shape-Guided Appearance-Based Generation.}
The proposed SA branch consists of an image encoder and a series of SA blocks.
The source image $I_s$ is first fed into the image encoder to produce the appearance code $F_0^I$, as shown in Figure~\ref{fig:method}.
The encoder consists of two convolutional layers in our experiments. 
The SA branch contains several cascaded SA blocks, which progressively update the initial appearance code $F_0^I$ to the final appearance code $F_T^I$ under the guidance of the AS branch. 
As can be seen from Figure~\ref{fig:method}, all SA blocks have identical network structures.
Consider the $t$-th block in Figure~\ref{fig:sa}, whose inputs are the appearance code $F_{t-1}^I {\in} \mathbb{R}^{c \times h \times w}$ and the shape code $F_{t-1}^P {\in} \mathbb{R}^{c \times h \times w}$. The output is the refined appearance code $F_t^I {\in} \mathbb{R}^{c \times h \times w}$.
Specifically, given the appearance code $F_{t-1}^I$, we first feed it into a convolutional layer to generate a new appearance code $C$, where $C {\in} \mathbb{R}^{c \times h \times w}$.
Then we reshape $C$ to $\mathbb{R}^{c \times (hw)}$, where $n {=} hw$ is the number of pixels.
At the same time, the SA block receives the shape code $F_{t-1}^P$ from the AS branch, which is also fed into a convolutional layer to produce a new shape code $B {\in} \mathbb{R}^{c \times h \times w}$. This is then reshaped to $\mathbb{R}^{c \times (hw)}$.
After that, we perform a matrix multiplication between the transpose of $C$ and $B$, and apply a Softmax layer to produce a correlation matrix $P {\in} \mathbb{R}^{(hw) \times (hw)}$:
\begin{equation}
\begin{aligned}
p_{ji} = {\rm Softmax} \left(\frac{{\rm exp} ( B_i  C_j)}{\sum_{i=1}^n {\rm exp}( B_i C_j)} \right),
\label{eq:non_local1}
\end{aligned}
\end{equation}
where $p_{ji}{\in}P$ measures the impact of the $i$-th position of $B$ on the $j$-th position of the appearance code $C$.
With this cross operation, the SA branch can capture more joint influence between the appearance code $F_{t-1}^I$ and shape code $F_{t-1}^P$, producing a richer appearance code~$F_{t}^I$.

\begin{figure*}[t] \small
	\centering
	\includegraphics[width=0.8\linewidth]{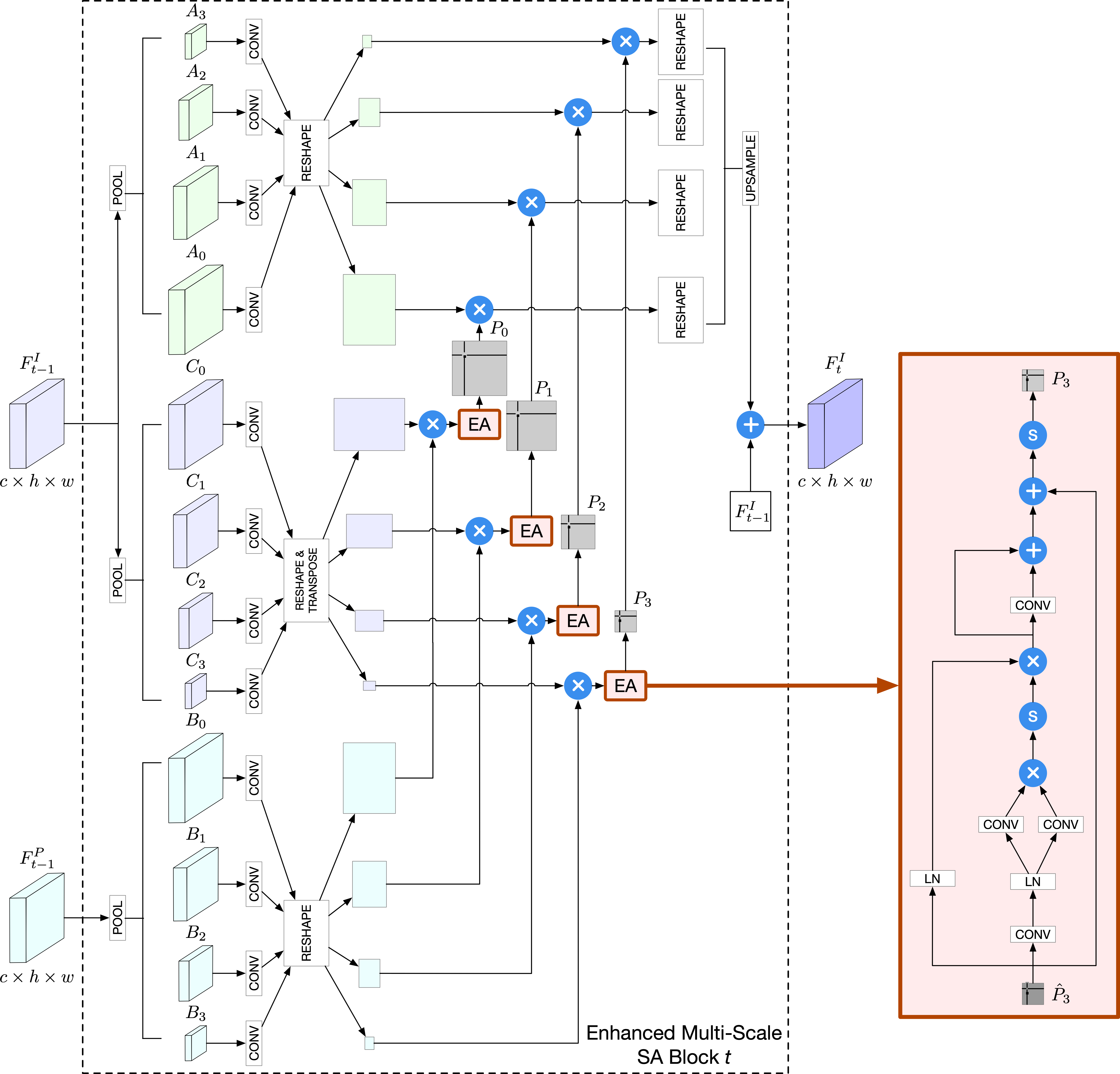}
	\caption{Structure of the proposed enhanced multi-scale SA block (proposed in this journal version), which takes the previous appearance code $F_{t-1}^I$ and the previous shape code $F_{t-1}^P$ as input and obtains the appearance code $F_{t}^I$ in a multi-scale crossed non-local way with our enhanced attention (EA) mechanism. The symbols $\oplus$, $\otimes$, and $\textcircled{s}$ denote element-wise addition, element-wise multiplication, and Softmax activation, respectively.}
	\label{fig:sa_ms}
	\vspace{-0.4cm}
\end{figure*}

Note that Equation~\eqref{eq:non_local1} is closely related to the non-local operator proposed by Wang et al. \cite{wang2018non}. 
The major difference is that the non-local operator in \cite{wang2018non} computes the pairwise similarity within the same feature map to incorporate global information, whereas the proposed cross operation computes the pairwise similarity between different feature maps, i.e., the person's appearance and shape feature maps.

After that, we feed $F_{t-1}^I$ into a convolutional layer to produce a new appearance code $A {\in} \mathbb{R}^{c \times h \times w}$ and reshape it to $\mathbb{R}^{c \times (h w)}$.
We then perform a matrix multiplication between $A$ and the transpose of $P$ and reshape the result to $\mathbb{R}^{c \times h \times w}$.
Finally, we multiply the result by a scale parameter $\alpha$ and conduct an element-wise sum operation with the original appearance code $F_{t-1}^I$ to obtain the refined appearance code $F_t^I {\in} \mathbb{R}^{c \times h \times w}$:
\begin{equation}
\begin{aligned}
F_t^I = \alpha \sum_{i=1}^{n}(p_{ji}  A_i) + F_{t-1}^I,
\end{aligned}
\end{equation}
where $\alpha$ is 0 in the beginning and is gradually updated. By doing so, each position of the refined appearance code $F_t^I$ is a weighted sum of all positions of the shape code $F_{t-1}^P$ and the previous appearance code $F_{t-1}^I$.
Thus, it has a global contextual view of $F_{t-1}^P$ and $F_{t-1}^I$, and it selectively aggregates useful context according to the correlation matrix~$P$. 

\noindent \textbf{Appearance-Guided Shape-Based Generation.}
In our preliminary experiments, we observed that using only the SA generation branch is not adequate for learning such a complex deformable translation process. Intuitively, since the shape features can guide the appearance features, we believe the appearance features can also be used to guide the shape features.
Therefore, we also propose an AS branch, which consists of a pose encoder and a sequence of AS blocks, as shown in Figure~\ref{fig:method}.
The source pose $P_s$ and target pose $P_t$ are first concatenated along the channel dimension and then fed into the pose encoder to produce the initial shape representation $F_0^P$.
The pose encoder has the same network structure as the image encoder.
Note that to capture the dependency between the two poses, we only adopt one pose encoder.

The AS branch contains several cascaded AS blocks, which progressively update the initial shape code $F_0^P$ to the final shape code $F_T^P$ under the guidance of the SA branch. 
All AS blocks have the same network structure, as illustrated in Figure~\ref{fig:method}.
Consider the $t$-th block, whose inputs are the shape code $F_{t-1}^P {\in} \mathbb{R}^{c \times h \times w}$ and the appearance code $F_{t-1}^I {\in} \mathbb{R}^{c \times h \times w}$. The output is the refined shape code $F_t^P {\in} \mathbb{R}^{c \times h \times w}$.

Specifically, given the shape code $F_{t-1}^P$, we first feed it into a convolutional layer to generate a new shape code $H$, where $H {\in} \mathbb{R}^{c \times h \times w}$. 
We then reshape $H$ to $\mathbb{R}^{c \times (h w)}$.
At the same time, the AS block receives the appearance code $F_{t-1}^I$ from the SA branch, which is also fed into a convolutional layer to produce a new appearance code $E$. This is then reshaped to $\mathbb{R}^{c \times (h w)}$.
After that, we perform a matrix multiplication between the transpose of $H$ and $E$, and apply a Softmax layer to produce another correlation matrix $Q {\in} \mathbb{R}^{(h w) \times (h w)}$:
\begin{equation}
\begin{aligned}
q_{ji} = {\rm Softmax} \left(\frac{{\rm exp} (E_i  H_j)}{\sum_{i=1}^n {\rm exp}(E_i  H_j)} \right),
\end{aligned}
\end{equation}
where $q_{ji}$ measures the impact of the $i$-th position of $E$ on the $j$-th position of the shape code $H$. $n {=} h w$ is the number of pixels.

Meanwhile, we feed $F_{t-1}^P$ into a convolutional layer to produce a new shape code $D {\in} \mathbb{R}^{c \times h \times w}$, which we then reshape to $\mathbb{R}^{c \times (h  w)}$.
We then perform a matrix multiplication between $D$ and the transpose of $Q$ and reshape the result to $\mathbb{R}^{c \times h \times w}$.
Finally, we multiply the result by a scale parameter $\beta$ and conduct an element-wise sum operation with the original shape code $F_{t-1}^P$.
The result is then concatenated with the appearance code $F_t^I$ and fed into a convolutional layer to obtain the updated shape code $F_t^P {\in} \mathbb{R}^{c \times h \times w}$:
\begin{equation}
\begin{aligned}
F_t^P = {\rm Concat}( \beta \sum_{i=1}^{n}(q_{ji} D_i) + F_{t-1}^P, F_t^I),
\end{aligned}
\end{equation}
where $\mathrm{Concat}(\cdot)$ denotes the channel-wise concatenation operation and $\beta$ is a parameter.
By doing so, each position in the refined shape code $F_t^P$ is a weighted sum of all positions in the appearance code $F_{t-1}^I$ and previous shape code $F_{t-1}^P$.

\noindent \textbf{Enhanced Multi-Scale Cross-Atteniton.}
Since there exists a large deformation between the source and the target poses, a single-scale cross-attention block may not be able to capture all the necessary spatial correlations for a fine-grained generation. Thus, we propose two enhanced multi-scale cross-attention blocks (i.e., EMSA and EMAS), obtaining multi-scale cross-correlations at different pose scales. 

Consider the $t$-th enhanced muti-scale SA (EMSA) block in Figure~\ref{fig:sa_ms}, whose inputs are the appearance code $F_{t-1}^I {\in} \mathbb{R}^{c \times h \times w}$ and the shape code $F_{t-1}^P {\in} \mathbb{R}^{c \times h \times w}$. The output is the refined appearance code $F_t^I {\in} \mathbb{R}^{c \times h \times w}$.
Specifically, given the appearance code $F_{t-1}^I$, we first feed it into a pyramid pooling module and a convolutional layer to generate new multi-scale appearance codes $[C_0, C_1, C_2, C_3]$, where $C_0 {\in} \mathbb{R}^{c \times \frac{h}{s_0} \times \frac{w}{s_0}}$, $C_1 {\in} \mathbb{R}^{c \times \frac{h}{s_1} \times \frac{w}{s_1}}$, $C_2 {\in} \mathbb{R}^{c \times \frac{h}{s_2} \times \frac{w}{s_2}}$, and $C_3 {\in} \mathbb{R}^{c \times \frac{h}{s_3} \times \frac{w}{s_3}}$.
Then we reshape $[C_0, C_1, C_2, C_3]$ to $\mathbb{R}^{c \times (\frac{hw}{s_0^2})}$, $\mathbb{R}^{c \times (\frac{hw}{s_1^2})}$, $\mathbb{R}^{c \times (\frac{hw}{s_2^2})}$, and $\mathbb{R}^{c \times (\frac{hw}{s_3^2})}$, respectively, where $n_0 {=} \frac{hw}{s_0^2}$, $n_1 {=} \frac{hw}{s_1^2}$, $n_2 {=} \frac{hw}{s_2^2}$, $n_3 {=} \frac{hw}{s_3^2}$ are the number of pixels.
Note that the number of pyramid levels and the size of each level can be modified. They are related to the size of the feature map input to the pyramid pooling layer. This structure abstracts different pose scales by employing pooling kernels of different strides. 
Our pyramid pooling module is a four-level module with bin sizes of $S_0{=}1$, $S_1{=}2$, $S_2{=}3$, and $S_3{=}6$, respectively.

At the same time, the EMSA block receives the shape code $F_{t-1}^P$, which is also fed into a  pyramid pooling module and a convolutional layer to produce new shape codes $[B_0, B_1, B_2, B_3]$, where $B_0 {\in} \mathbb{R}^{c \times \frac{h}{s_0} \times \frac{w}{s_0}}$, $ B_1 {\in} \mathbb{R}^{c \times \frac{h}{s_1} \times \frac{w}{s_1}}$, $B_2 {\in} \mathbb{R}^{c \times \frac{h}{s_2} \times \frac{w}{s_2}}$, and $B_3 {\in} \mathbb{R}^{c \times \frac{h}{s_3} \times \frac{w}{s_3}}$. These four codes are then reshaped to $\mathbb{R}^{c \times (\frac{hw}{s_0^2})}$, $\mathbb{R}^{c \times (\frac{hw}{s_1^2})}$, $\mathbb{R}^{c \times (\frac{hw}{s_2^2})}$, and $\mathbb{R}^{c \times (\frac{hw}{s_3^2})}$, respectively.
After that, we perform a matrix multiplication between the transpose of $[C_0, C_1, C_2, C_3]$ and $[B_0, B_1, B_2, B_3]$, and apply the proposed enhanced attention (EA) to produce four correlation matrices $[P_0, P_1, P_2, P_3]$, where $P_0 {\in} \mathbb{R}^{(\frac{hw}{s_0^2}) \times (\frac{hw}{s_0^2})}$, $P_1 {\in} \mathbb{R}^{(\frac{hw}{s_1^2}) \times (\frac{hw}{s_1^2})}$, $P_2 {\in} \mathbb{R}^{(\frac{hw}{s_2^2}) \times (\frac{hw}{s_2^2})}$, and $P_3 {\in} \mathbb{R}^{(\frac{hw}{s_3^2}) \times (\frac{hw}{s_3^2})}$:
\begin{equation}
\begin{aligned}
{p_0}_{ji} = \frac{{\rm exp} ( {B_0}_i  {C_0}_j)}{\sum_{i=1}^n {\rm exp}( {B_0}_i {C_0}_j)}, \quad
{p_1}_{ji} = \frac{{\rm exp} ( {B_1}_i  {C_1}_j)}{\sum_{i=1}^n {\rm exp}( {B_1}_i {C_1}_j)}, \\
{p_2}_{ji} = \frac{{\rm exp} ( {B_2}_i  {C_2}_j)}{\sum_{i=1}^n {\rm exp}( {B_2}_i {C_2}_j)}, \quad
{p_3}_{ji} = \frac{{\rm exp} ( {B_3}_i  {C_3}_j)}{\sum_{i=1}^n {\rm exp}( {B_3}_i {C_3}_j)},
\label{eq:non_local2}
\end{aligned}
\end{equation}
where ${p_0}_{ji}{\in}P_0$, ${p_1}_{ji}{\in}P_1$, ${p_2}_{ji}{\in}P_2$, ${p_3}_{ji}{\in}P_3$ measure the impact of the $i$-th position of $B_0$, $B_1$, $B_2$, $B_3$ on the $j$-th position of the corresponding appearance code $C_0$, $C_1$, $C_2$, $C_3$.
With this cross operation, the SA branch can capture more joint influence between the appearance code $F_{t-1}^I$ and shape code $F_{t-1}^P$ at different pose scales, producing a richer appearance code~$F_{t}^I$.

After that, we feed $F_{t-1}^I$ into anthor pyramid pooling module and another convolutional layer to produce new appearance codes $[A_0, A_1, A_2, A_3]$, where $A_0 {\in} \mathbb{R}^{c \times \frac{h}{s_0} \times \frac{w}{s_0}}$, $A_1 {\in} \mathbb{R}^{c \times \frac{h}{s_1} \times \frac{w}{s_1}}$, $A_2 {\in} \mathbb{R}^{c \times \frac{h}{s_2} \times \frac{w}{s_2}}$, $A_3 {\in} \mathbb{R}^{c \times \frac{h}{s_3} \times \frac{w}{s_3}}$, and reshape those to $\mathbb{R}^{c \times (\frac{hw}{s_0^2})}$, $\mathbb{R}^{c \times (\frac{hw}{s_1^2})}$, $\mathbb{R}^{c \times (\frac{hw}{s_2^2})}$, and $\mathbb{R}^{c \times (\frac{hw}{s_3^2})}$, respectively.
We then perform a matrix multiplication between $[A_0, A_1, A_2, A_3]$ and the transpose of $[P_0, P_1, P_2, P_3]$ and reshape the results to $\mathbb{R}^{c \times \frac{h}{s_0} \times \frac{w}{s_0}}$, $\mathbb{R}^{c \times \frac{h}{s_1} \times \frac{w}{s_1}}$, $\mathbb{R}^{c \times \frac{h}{s_2} \times \frac{w}{s_2}}$, and $\mathbb{R}^{c \times \frac{h}{s_3} \times \frac{w}{s_3}}$, respectivley.
Then we upsample the low-dimension feature maps to get the same size feature as the original feature map via bilinear interpolation.
Thus, different levels of features are concatenated as the
final pyramid pooling global feature.
Finally, we multiply the result by a scale parameter $\alpha$ and conduct an element-wise sum operation with the original appearance code $F_{t-1}^I$ to obtain the refined appearance code $F_t^I {\in} \mathbb{R}^{c \times h \times w}$:
\begin{equation}
\begin{aligned}
F_t^I = \alpha {\rm Conv}\{ {\rm Concat} [{\rm Up}(\sum_{i=1}^{n_0}{p_0}_{ji}  {A_0}_i), {\rm Up}(\sum_{i=1}^{n_1}{p_1}_{ji}  {A_1}_i), \\ {\rm Up}(\sum_{i=1}^{n_2}{p_2}_{ji}  {A_2}_i), {\rm Up}(\sum_{i=1}^{n_3}{p_3}_{ji}  {A_3}_i)]\} + F_{t-1}^I,
\end{aligned}
\end{equation}
where $\alpha$ is 0 in the beginning and is gradually updated;
$\mathrm{Conv}(\cdot)$ is a convolutional layer; 
$\mathrm{Concat}(\cdot)$ is a function for channel-wise concatenation operation; $\mathrm{Up}(\cdot)$ is an upsampling operation.
By doing so, each position of the refined appearance code $F_t^I$ is a weighted sum of all positions of the shape code $F_{t-1}^P$ and the previous appearance code $F_{t-1}^I$ at different pose scales.

At the same time, the enhanced multi-scale AS (EMAS) block has the same network structure as the EMSA block.

\noindent \textbf{Enhanced Attention.}
In traditional attention blocks, the correlation calculation between each query-key pair in the correlation maps $[P_0, P_1, P_2, P_3]$ in Equation \eqref{eq:non_local2} is performed individually, disregarding the correlations among other query-key pairs. This independent correlation calculation method can generate erroneous correlations caused by imperfect feature representations or the existence of distracting image patches within cluttered background scenes. These inaccurate correlations can potentially introduce noisy and ambiguous attention, hindering effective information propagation during cross-attention and resulting in suboptimal performance.

To tackle the mentioned challenge, we present a new module called the enhanced attention (EA) module designed to enhance the accuracy of the correlation maps $[P_0, P_1, P_2, P_3]$. In typical scenarios, when a key displays a strong correlation with a query, its neighboring keys also tend to exhibit relatively high correlations with the same query. Conversely, correlations that deviate from this pattern are likely to be noise. Drawing inspiration from this observation, we introduce the EA module, which utilizes the correlation information within $[P_0, P_1, P_2, P_3]$ to improve correlation consistency. By focusing on achieving correlation consistency around each key, our proposed EA module aims to amplify genuine correlations between relevant query-key pairs while suppressing false correlations among unrelated pairs.

To refine the correlation maps $[P_0, P_1, P_2, P_3]$, we introduce an additional EA module, depicted in Figure \ref{fig:sa_ms} (right). This EA module is a modified version of traditional attention. We consider the columns within the correlation maps $[P_0, P_1, P_2, P_3]$ as a sequential arrangement of correlation vectors. The EA module utilizes these correlation vectors as queries, keys, and values to generate a residual correlation map as the output.

Take Figure \ref{fig:sa_ms} (right) as an example. We can obtain $\hat{P}_3$ from the above multi-scale cross-attention module.
Next, $\hat{P}_3$ first passes through a convolution layer to reduce the dimension to $n$, then passes through a LayerNorm layer to normalize and then passes through two convolution layers to obtain the key $K$ and query $Q$, respectively. At the same time, $\hat{P}_3$ gets value $V$ through another LayerNorm layer.
This process can be expressed as:
\begin{equation}
\begin{aligned}
EA_1(\hat{P}_3) & = {\rm Softmax} \left(\frac{{\rm exp} (K_i  Q_j)}{\sum_{i=1}^n {\rm exp}(K_i  Q_j)} \right) V_i, \\
EA_2(\hat{P}_3) & = EA_1(\hat{P}_3) + {\rm Conv}(EA_1(\hat{P}_3)).
\end{aligned}
\end{equation}
The EA module produces a residual correlation vector for each correlation vector in the correlation map $\hat{P}_3$ by aggregating the original correlation vectors. This process can be seen as seeking consensus among correlations within a global receptive field. Utilizing the residual correlation map, the final correlation map $P_3$ can be obtained as follows:
\begin{equation}
\begin{aligned}
P_3  = {\rm Softmax} \left( EA_2(\hat{P}_3) + \hat{P}_3 \right).
\end{aligned}
\end{equation}
For other scale feature maps in Equation \eqref{eq:non_local2}, we obtain the corresponding correlation maps $[P_0, P_1, P_2]$ similarly.

\noindent \textbf{Densely Connected Co-Attention Fusion.}
The proposed DCCAF module is inspired by DenseNet \cite{huang2017densely} and consists of two steps, i.e., generating intermediate results and co-attention maps.
The co-attention maps are used to spatially select from both the intermediate generations and the input image and are combined to synthesize a final output.
The proposed module is different from the multi-channel attention selection module in SelectionGAN \cite{tang2019multi}: (1) We use two generation branches to generate intermediate results, i.e., the SA branch and the AS branch.
(2) Attention maps are generated by the combination of both shape and appearance features, so the model learns more correlations between the two.
(3) We consider the shape and appearance features from all the blocks to produce the final feature representation.
(4) We also produce the input attention map, which aims to select useful content from the input image for generating the final image.

We consider two directions to generate intermediate results.
One is generating multiple intermediate images from  appearance codes $[F_0^I, \cdots F_{t-1}^I, F_t^I, \cdots F_T^I]$, and the other is generating them from  shape codes $[F_0^P, \cdots F_{t-1}^P, F_t^P, \cdots F_T^P]$.
Specifically, the appearance codes $[F_0^I, \cdots F_{t-1}^I, F_t^I, \cdots F_T^I]$ is concatenated and fed into a decoder to generate $N$ intermediate results $I^I{=}\{I_i^I\}_{i=1}^N$, followed by a $\rm{Tanh}$ activation function.
Meanwhile, the shape codes $[F_0^P, \cdots F_{t-1}^P, F_t^P, \cdots F_T^P]$ is concatenated and fed into another decoder to generate another $N$ intermediate results $I^P{=}\{I_i^P\}_{i=1}^{N}$. This is also followed by a $\rm{Tanh}$ activation function.
Both can be formulated as,
\begin{equation}
\begin{aligned}
I_i^I = {}& {\rm Tanh}({\rm Concat}(F_0^I, \cdots F_{t-1}^I, F_t^I, \cdots F_T^I) W_i^I +b_i^I), \\
I_i^P ={} & {\rm Tanh}({\rm Concat}(F_0^P, \cdots F_{t-1}^P, F_t^P, \cdots F_T^P) W_i^P +b_i^P), 
\end{aligned}
\end{equation}
where $i {=} 1, {\cdots}, N$ and two convolution operations are performed with $N$ convolutional filters $\{W_i^I, b_i^I\}_{i=1}^{N}$ and $\{W_i^P, b_i^P\}_{i=1}^{N}$.
Thus, the $2N$ intermediate results and the input image $I_s$ can be regarded as the candidate image pool.

\begin{table*}[!t] \small
	\centering
	\caption{Quantitative results on Market-1501 and DeepFashion. ($\ast$) denotes the results on our test set.}
		\begin{tabular}{rcccccccc} \toprule
			\multirow{2}{*}{Method}  & \multicolumn{5}{c}{Market-1501} & \multicolumn{3}{c}{DeepFashion} \\ \cmidrule(lr){2-6} \cmidrule(lr){7-9} 
			 & SSIM $\uparrow$ & IS $\uparrow$  & Mask-SSIM $\uparrow$ & Mask-IS $\uparrow$  & PCKh $\uparrow$ & SSIM $\uparrow$ & IS $\uparrow$  & PCKh $\uparrow$ \\ \hline	
			PG2~\cite{ma2017pose}                                        & 0.253 & 3.460 & 0.792 & 3.435   & - & 0.762 & 3.090  & - \\
			DPIG~\cite{ma2018disentangled}                           & 0.099 & 3.483 & 0.614 & 3.491   & - & 0.614 & 3.228    & - \\
			PoseGAN~\cite{siarohin2018deformable}              & 0.290 & 3.185 & 0.805 & 3.502   & - & 0.756 & 3.439    & -\\ 
			C2GAN~\cite{tang2019cycle}                                & 0.282 & 3.349 & 0.811 & 3.510   & - & -     & -            & -\\ 
			BTF~\cite{albahar2019guided}                               & -     & -     & -     & -       & - & 0.767 & 3.220                   & -\\ \hline
			PG2$^\ast$~\cite{ma2017pose}                             & 0.261 & 3.495 & 0.782 & 3.367   &0.73 & 0.773 & 3.163  & 0.89 \\ 
			PoseGAN$^\ast$~\cite{siarohin2018deformable}    & 0.291 & 3.230 & 0.807 & 3.502   & 0.94 & 0.760 & 3.362 & 0.94 \\ 
			VUnet$^\ast$~\cite{esser2018variational}              & 0.266 & 2.965 & 0.793 & 3.549   & 0.92 & 0.763 & 3.440 & 0.93 \\
			PoseWarp$^\ast$~\cite{balakrishnan2018synthesizing} &  - & - & - & - & - & 0.764 & 3.368 & 0.93 \\
			CMA$^\ast$~\cite{chi2019two} &  - & - & - & - & - & 0.768 & 3.213 & 0.92 \\
            PoT-GAN$^\ast$ \cite{li2021pot} & - & - & - & - & - & 0.775 & 3.365 & - \\
            SCMNet$^\ast$ \cite{wang2022self} & - & - & - & - & - & 0.751 & \textbf{3.632} & - \\
            CASD$^\ast$ \cite{zhou2022cross} & - & - & - & - & - & 0.725 &  - & - \\
            DBT$^\ast$ \cite{zhang2022exploring} & 0.285 & - & -& - & - & 0.778 & - &-\\
   		PATN$^\ast$~\cite{zhu2019progressive}  & 0.311 & 3.323 & 
             0.811 & 3.773   & 0.94 & 0.773 & 3.209 & 0.96 \\ 
             BiGraphGAN$^\ast$ \cite{tang2020bipartite} & 0.325 &  3.329 & 0.818 & 3.695 & 0.94 & 0.778 & 3.430 & 0.97  \\ 
            PoseStylizer$^\ast$~\cite{huang2020generating} & 0.312 & 
            3.132 & 0.808 & 3.729 & 0.94 & 0.775 & 3.295 & 0.96 \\ 
            SPGNet$^\ast$ \cite{lv2021learning} & 0.315 & - & 0.818 & - & 0.97 & 0.782 & - & 0.97 \\ 
            EPIG$^\ast$ \cite{shen2022exploiting} & 0.322 & 3.318 & 0.816 & 3.780 & 0.94 & 0.773 & 3.216 & 0.96 \\
            MSA$^\ast$ \cite{roy2023multi} & 0.266 & \textbf{3.682} & - & - & 0.95 & 0.769  & 3.379 & \textbf{0.98} \\
            \hline
		XingGAN \cite{tang2020xinggan} & 0.313 & 3.506 & 
            0.816 & 3.872 & 0.93 & 0.778 & 3.476  & 0.95 \\ 
            XingGAN++ (Ours) & \textbf{0.333} & 3.645 & \textbf{0.828} & \textbf{3.903} & \textbf{0.97} & \textbf{0.803} & 3.524 & \textbf{0.98} \\ \hline	
			Real Data                                                               & 1.000 & 3.890 & 1.000 & 3.706   & 1.00 & 1.000 & 4.053 & 1.00 \\	
			\bottomrule	
	\end{tabular}
	\label{tab:pose_reuslts}
	\vspace{-0.4cm}
\end{table*}

To generate co-attention maps that reflect the correlation between the appearance $F^I{=}[F_0^I, \cdots F_{t-1}^I, F_t^I, \cdots F_T^I]$ and shape $F^P{=} [F_0^P, \cdots F_{t-1}^P, F_t^P, \cdots F_T^P]$ codes, we first stack both $F^I$ and $F^P$ along the channel axis, and then feed them into a group of filters $\{W_i^A, b_i^A\}_{i=1}^{2N+1}$ to generate the corresponding $2N{+}1$ co-attention maps:
\begin{equation}
\begin{aligned}
I_i^A =  {\rm Softmax}({\rm Concat}(F^I, F^P) W_i^A +b_i^A),  \\   {\rm for}~i  =  1, \cdots, 2N{+}1 
\end{aligned}
\end{equation}
where ${\rm Softmax}$ is a channel-wise Softmax function used for normalization, and $\mathrm{Concat}(\cdot)$ denotes the channel-wise concatenation operation.
Finally, the learned co-attention maps are used to perform a channel-wise content selection from each intermediate generation and the input image, as follows:
\begin{equation}
\begin{aligned}
I_t^{'} = (I_1^A \otimes I_1^I) \oplus \cdots (I_{2N}^A \otimes I_{2N}^P)  \oplus (I_{2N+1}^A \otimes I_s),
\end{aligned}
\end{equation}
where $I_t^{'}$ represents the final synthesized person image selected from the multiple diverse results and the input image. $\otimes$ and $\oplus$ denote the element-wise multiplication and addition, respectively.

\noindent \textbf{Optimization Objective.}
We use three different losses as our full optimization objective, i.e., the adversarial loss $\mathcal{L}_{gan}$, pixel loss $\mathcal{L}_{l1}$, and perceptual loss $\mathcal{L}_{p}$:
\begin{equation}
\begin{aligned}
\min_G \max_{D_I, D_P}  \mathcal{L} = \lambda_{gan} \mathcal{L}_{gan} + \lambda_{l1} \mathcal{L}_{l1} + \lambda_{p} \mathcal{L}_{p},
\label{eq:loss} 
\end{aligned}
\end{equation}
where $\lambda_{gan}$, $\lambda_{l1}$ and $\lambda_{p}$ are the weights, which measure the corresponding contributions of each loss to the overall loss $\mathcal{L}$.
The overall adversarial loss is derived from the appearance-guided discriminator $D_I$ and the shape-guided discriminator $D_P$, which aim to determine how likely it is that $I_t^{'}$ contains the same person as $I_s$ and how well $I_t^{'}$ aligns with the target pose $P_t$, respectively.
The $L_1$ pixel loss is used to compute the difference between the generated image $I_t^{'}$ and the real target image $I_t$, i.e., $\mathcal{L}_{l1}{=} \vert\vert I_t-I_t^{'}\vert\vert_1$.
The perceptual loss $\mathcal{L}_{p}$ is used to reduce pose distortions and make the generated images look more natural and smooth, i.e., $\mathcal{L}_{p}{=}\vert\vert \phi(I_t)-\phi(I_t^{'})\vert\vert_1$, where $\phi$ denotes the outputs of several layers in the pre-trained VGG19 network~\cite{simonyan2014very}.

\noindent \textbf{Implementation Details.}
We follow the standard training procedure of GANs and alternately train the proposed generator $G$ and two discriminators ($D_I$, $D_P$).
During training, $G$ takes $I_s$, $P_s$ and $P_t$ as input and outputs a translated person image $I_t^{'}$ with target pose $P_t$.
Specifically, $I_s$ is fed to the SA branch, and $P_s$, $P_t$ are fed to the AS branch.
For the adversarial training, ($I_s$, $I_t$) and ($I_s$, $I_t^{'}$) are fed to the appearance-guided discriminator $D_P$ to ensure appearance consistency.
($P_t$, $I_t$) and ($P_t$, $I_t^{'}$) are fed to the shape-guided discriminator $D_P$ to ensure shape consistency.

The Adam optimizer \cite{kingma2014adam} is used to train the proposed methods for around 90K iterations with $\beta_1{=}0.5$ and $\beta_2{=}0.999$.
We set $T{=}9$ in the proposed Xing generator, $T{=}5$ in the proposed Xing++ generator, and $N{=}10$ in the proposed densely connected co-attention fusion module on both datasets used.
After grid-search experiments, $\lambda_{gan}$, $\lambda_{l1}$ and $\lambda_{p}$ in Equation~\eqref{eq:loss} are set to 5, 50 and 50, respectively.
For the decoders, the kernel size of convolutions for generating the intermediate images and co-attention maps are $3 {\times} 3$ and $1 {\times} 1$, respectively.

%% file: 4experiments.tex
\section{Experiments}

In this section, we present experimental results and analyses of the proposed XingGAN++ on the challenging person image generation task.

\begin{figure*}[!ht]\small
	\centering
	\subfigure[]{\label{fig:market1}\includegraphics[width=0.68\linewidth]{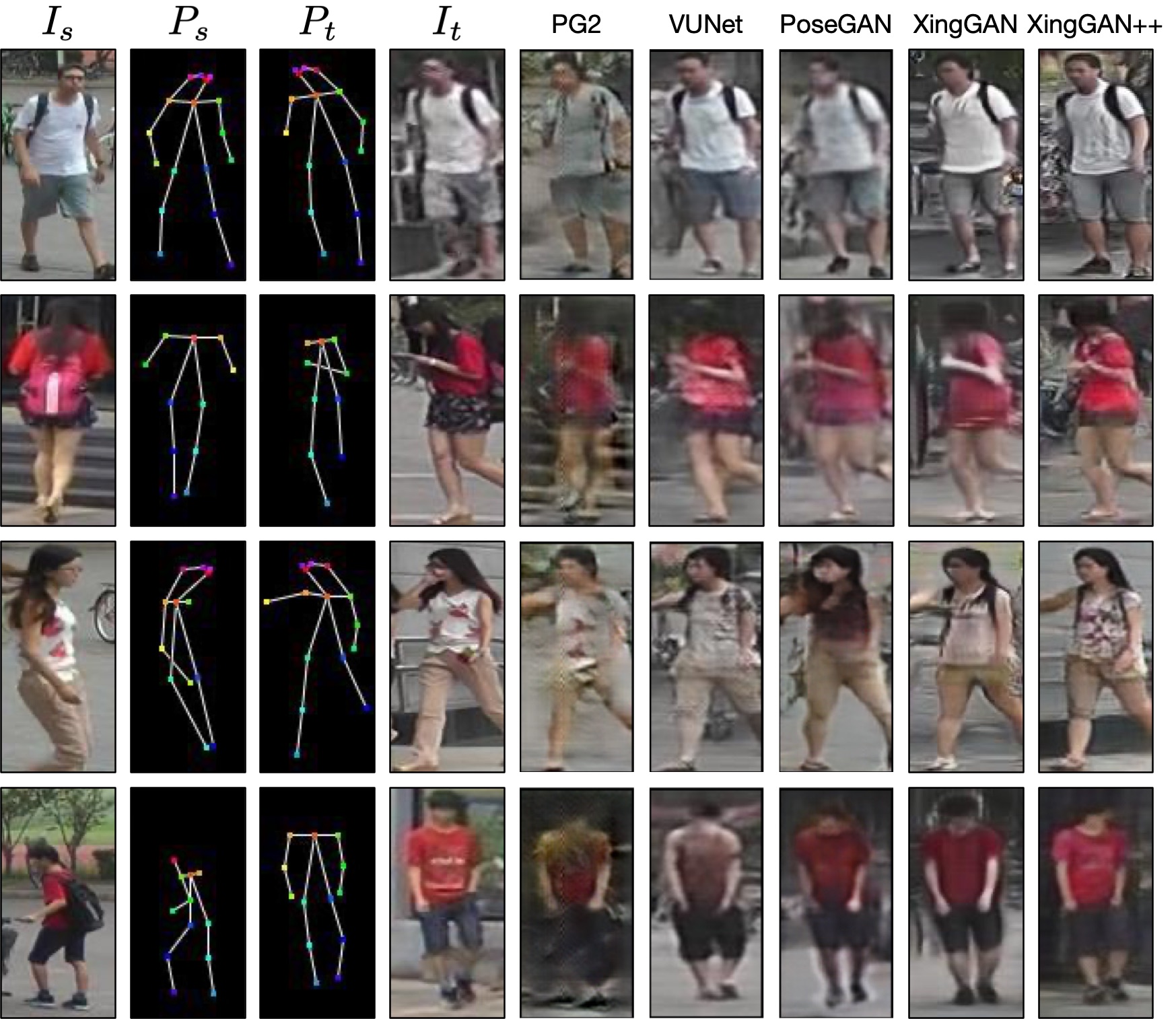}}
	\subfigure[]{\label{fig:market2}\includegraphics[width=0.75\linewidth]{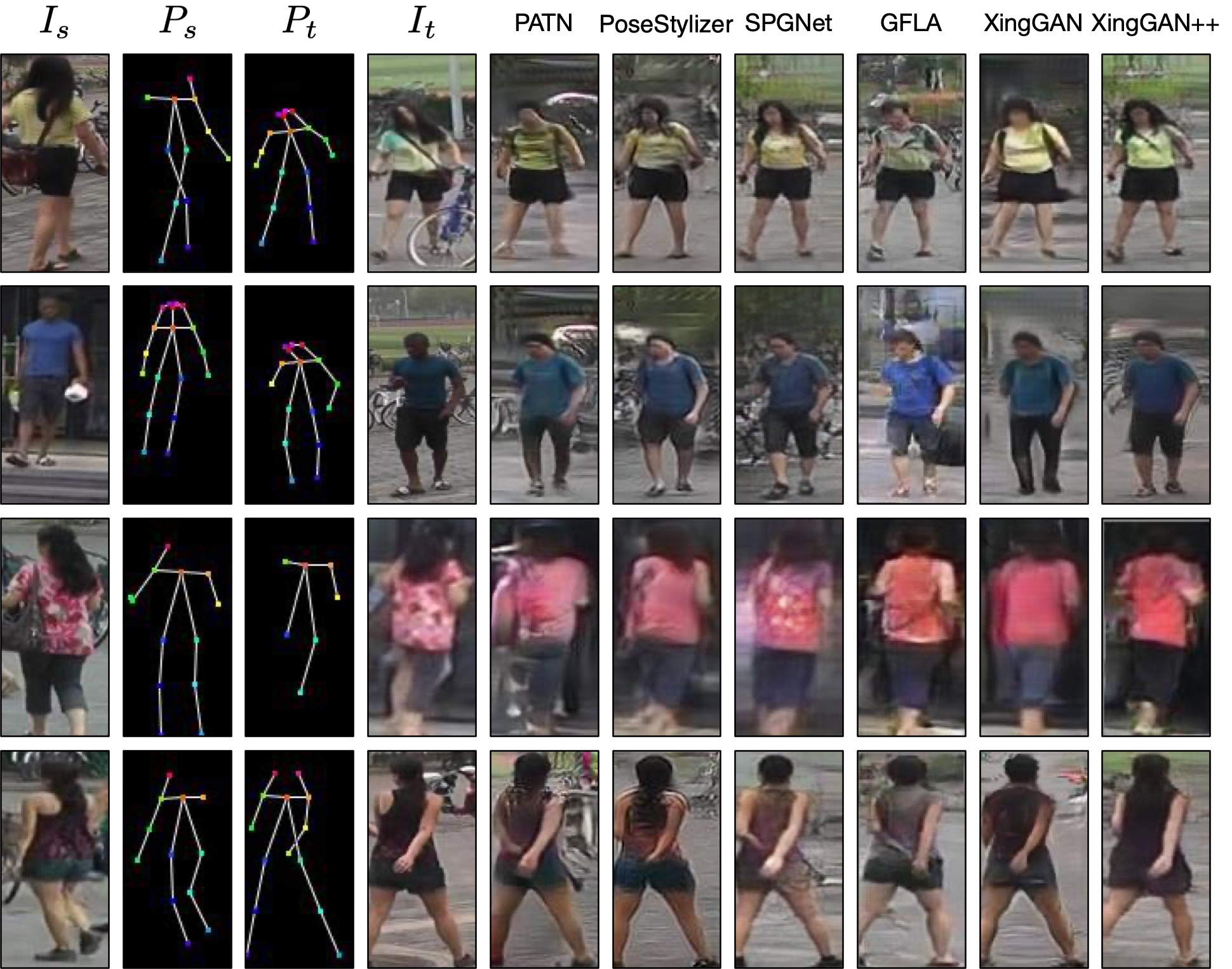}}
	\caption{Qualitative comparisons  on Market-1501. (a) From left to right: Source Image ($I_s$), Source Pose ($P_s$), Target Pose ($P_t$), Target Image ($I_t$), PG2~\cite{ma2017pose}, VUNet~\cite{esser2018variational}, PoseGAN~\cite{siarohin2018deformable}, XingGAN \cite{tang2020xinggan}, and XingGAN++ (Ours). (b) From left to right: Source Image ($I_s$), Source Pose ($P_s$), Target Pose ($P_t$), Target Image ($I_t$), PATN~\cite{zhu2019progressive}, PoseStylizer~\cite{huang2020generating}, SPGNet \cite{lv2021learning}, GFLA \cite{ren2020deep}, XingGAN \cite{tang2020xinggan}, and XingGAN++ (Ours).}
	\label{fig:market_result}
	\vspace{-0.4cm}
\end{figure*}

\begin{figure*}[!htbp]\small
	\centering
	\subfigure[]{\label{fig:fashion1}\includegraphics[width=0.7\linewidth]{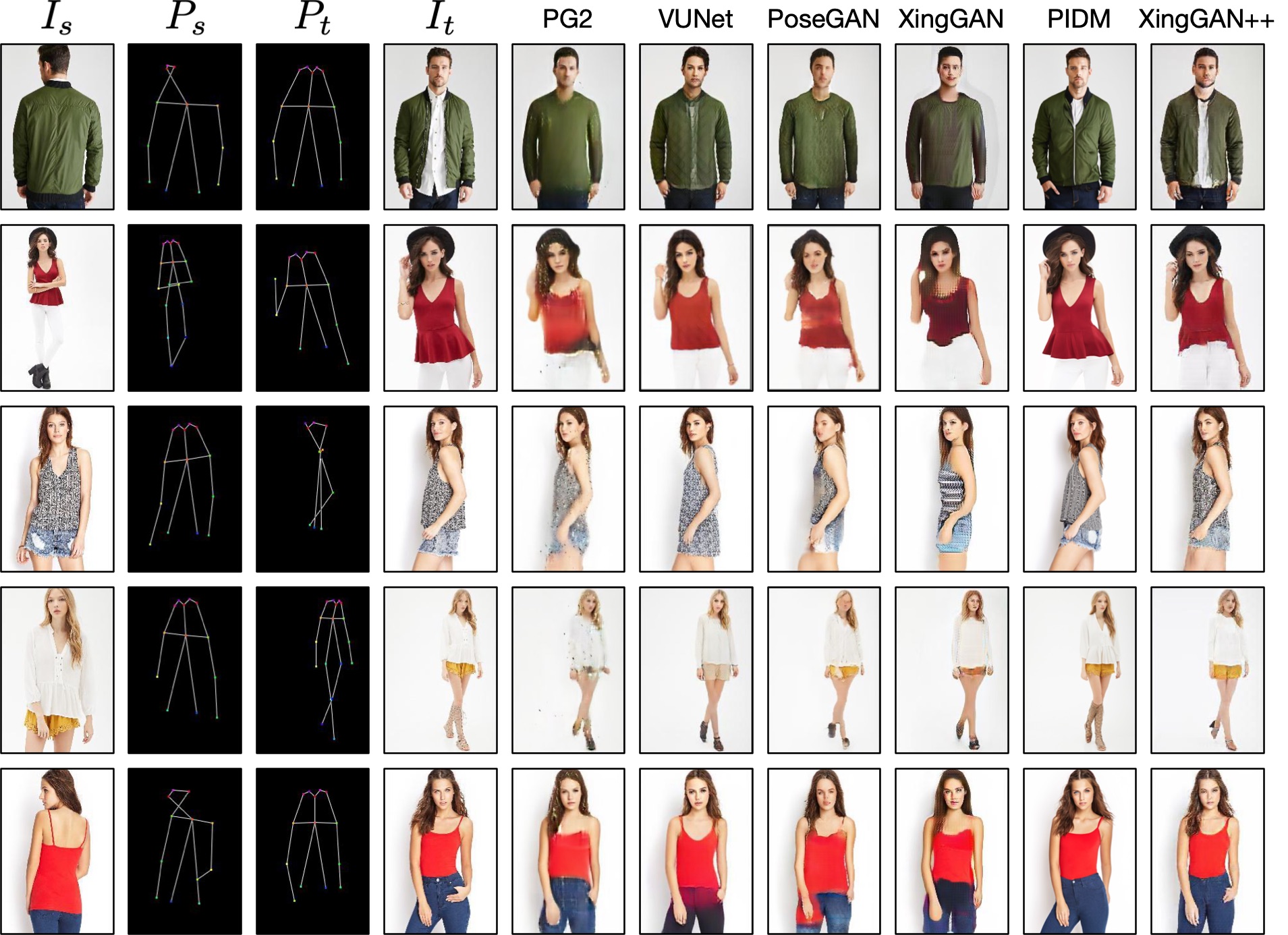}}
	\subfigure[]{\label{fig:fashion2}\includegraphics[width=0.77\linewidth]{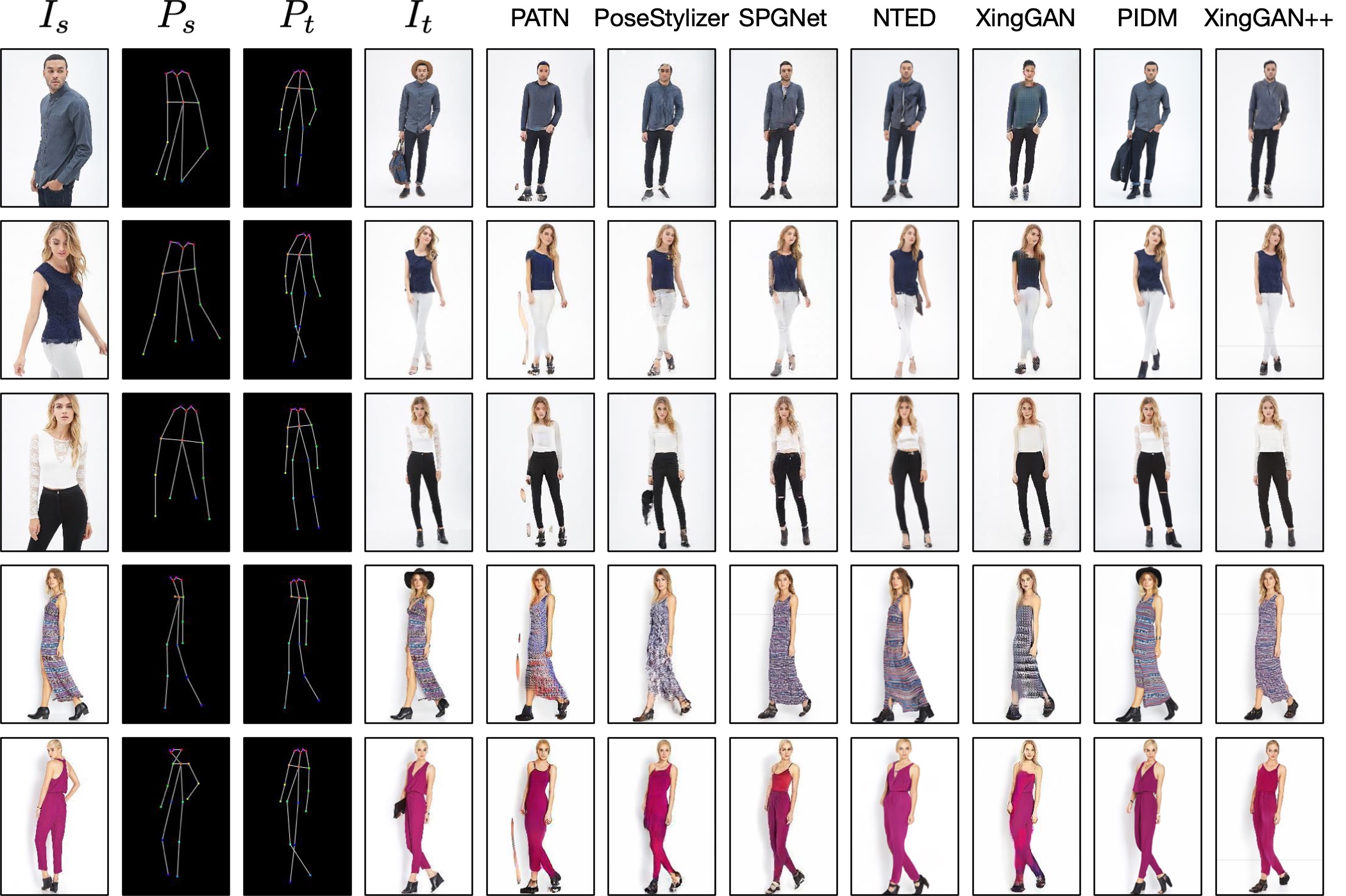}}
	\caption{Qualitative comparisons of person pose generation on DeepFashion. (a) From left to right: Source Image ($I_s$), Source Pose ($P_s$), Target Pose ($P_t$), Target Image ($I_t$), PG2~\cite{ma2017pose}, VUNet~\cite{esser2018variational}, PoseGAN~\cite{siarohin2018deformable}, XingGAN \cite{tang2020xinggan}, and XingGAN++ (Ours). (b) From left to right: Source Image ($I_s$), Source Pose ($P_s$), Target Pose ($P_t$), Target Image ($I_t$), PATN~\cite{zhu2019progressive}, PoseStylizer~\cite{huang2020generating}, SPGNet \cite{lv2021learning}, NTED \cite{ren2022neural}, XingGAN \cite{tang2020xinggan}, PIDM \cite{bhunia2023person}, and XingGAN++ (Ours).}
	\label{fig:fashion_result}
	\vspace{-0.4cm}
\end{figure*}

\subsection{Experimental Setups}
\noindent \textbf{Datasets.}
We follow \cite{ma2017pose,siarohin2018deformable,zhu2019progressive} and conduct experiments on two challenging datasets, i.e., Market-1501 \cite{zheng2015scalable} and DeepFashion \cite{liu2016deepfashion}.
The images from Market-1501 and DeepFashion are rescaled to $128 {\times} 64$ and $256 {\times} 256$, respectively.
To generate human skeletons as training data, we employ OpenPose \cite{cao2017realtime} to extract human joints. 
In this way, both $P_s$ and $P_t$ consist of an 18-channel heat map encoding the positions of 18 joints of a human body.
We also filter out images in which no human is detected.
Thus, we collected 101,966 training pairs and 8,570 testing pairs for DeepFashion.
For Market-1501, we have 263,632 training and 12,000 testing pairs. 
Note that, to better evaluate the proposed methods, the person identities of the training set do not overlap with those of the test set.

\noindent \textbf{Evaluation Metrics.}
We follow \cite{ma2017pose,siarohin2018deformable,zhu2019progressive,tang2020bipartite} and adopt structural similarity index measure (SSIM) \cite{wang2004image}, Inception score (IS) \cite{salimans2016improved}, and their masked versions, i.e., Mask-SSIM and Mask-IS, as our evaluation metrics.
Moreover, we adopt the PCKh score proposed in \cite{zhu2019progressive} to explicitly assess the shape consistency.

\begin{table}[!t] \small
	\centering
	\caption{Quantitative results on DeepFashion using $256 {\times} 176$ images.}
		\begin{tabular}{rcccccc} \toprule
			Method & SSIM $\uparrow$ & LPIPS $\downarrow$ & FID $\downarrow$ \\ \midrule
			PATN~\cite{zhu2019progressive}  & 0.6714 &0.2533 & 20.728  \\
                ADGAN \cite{men2020controllable} & 0.6735 & 0.2255 & 14.540 \\
                PISE \cite{zhang2021pise} & 0.6537 & 0.2244 & 11.518\\
                GFLA \cite{ren2020deep}  & 0.7082 & 0.1878 & 9.8272 \\
                NTED \cite{ren2022neural} & 0.7182 & 0.1752 & 8.6838\\
			XingGAN \cite{tang2020xinggan} & 0.7042& 0.1823 &8.9621 \\	
                PIDM \cite{bhunia2023person} & \textbf{0.7312} & 0.1678 & \textbf{6.3671} \\
                XingGAN++ (Ours) & 0.7262 & \textbf{0.1645} & 7.6260 \\
			\bottomrule	
	\end{tabular}
	\label{tab:pose_nted}
	\vspace{-0.3cm}
\end{table}

\begin{table}[!t] \small
	\centering
	\caption{Quantitative results on DeepFashion using $512 {\times} 352$ images and DensePose as an additional condition.}
		\begin{tabular}{rcccccc} \toprule
			Method & SSIM $\uparrow$ & LPIPS $\downarrow$ & FID $\downarrow$ \\ \midrule
			CocosNet2 \cite{zhou2021cocosnet} & 0.7236 & 0.2265 & 13.325 \\
                NTED \cite{ren2022neural} & 0.7376 & 0.1980 & 7.7821\\
			PoCoLD \cite{han2023controllable} & 0.7430 & 0.1920  & 8.4163 \\	
                PIDM \cite{bhunia2023person} & 0.7419 & \textbf{0.1768} & \textbf{5.8365} \\
                XingGAN++ (Ours) & \textbf{0.7482} & 0.1834 & 7.6583 \\
			\bottomrule	
	\end{tabular}
	\label{tab:densepose}
	\vspace{-0.4cm}
\end{table}

\subsection{Experimental Results}
\noindent \textbf{Quantitative Comparisons.}
We compare the proposed XingGAN++ with several leading methods in Table~\ref{tab:pose_reuslts}.
Quantitative results measured by SSIM, IS, Mask-SSIM, Mask-IS, and PCKh.
Note that previous works \cite{ma2017pose,siarohin2018deformable} have not released their train/test splits, so we use their previously trained models and re-evaluate their performance on our test set, as in PATN~\cite{zhu2019progressive}.
Although our test set inevitably includes some of their training samples, the proposed XingGAN++ achieves the best results in terms of all the metrics on both datasets (except IS), validating the effectiveness of the proposed method.
Our method achieves the highest SSIM score, which highlights strong perceptual similarity with real images. However, the IS is not the best, likely because SSIM and IS tend to exhibit a negative correlation. SSIM focuses on structural similarity and perceptual quality, while IS emphasizes diversity and semantic clarity. This difference can result in a high SSIM but relatively lower IS. 

Moreover, we follow the experimental setting in NTED \cite{ren2022neural} and train our model using the ${256 \times}176$ images proposed in NETD. The results of SSIM, LPIPS \cite{zhang2018unreasonable}, and FID \cite{heusel2017gans} are shown in Table \ref{tab:pose_nted}, we see that the proposed method still achieves competitive results in all three metrics.
We also follow the approach outlined in PoCoLD \cite{han2023controllable} and incorporate DensePose as an additional condition to validate the effectiveness of our method comprehensively. 
DensePose provides significantly richer and more detailed body structure information compared to traditional human skeleton representations. 
The results compared with SOTA methods are shown in Table~\ref{tab:densepose}.
We see that the proposed method still achieves competitive results, validating its effectiveness.

Diffusion-based method, i.e., PIDM, achieves the best results on most metrics in Tables \ref{tab:pose_nted} and \ref{tab:densepose}. This is because PIDM progressively refines the generated image through a series of iterative steps, allowing it to gradually approach the final target distribution with high accuracy. This process helps diffusion models produce images with superior quality, particularly in terms of structural coherence and fidelity. However, this iterative refinement is computationally demanding, requiring significantly more time and resources compared to our GAN-based method, which generates images in a single forward pass and are therefore more efficient in terms of speed.
We observe that our method achieves competitive results compared to PIDM. However, our method is 69.33$\times$ faster than PIDM during training and 18.04$\times$ faster during inference, as shown in Table \ref{tab:time}.

\noindent \textbf{Data Augmentation for Person Re-Identification.}
We also conduct person re-identification experiments using the augmented data generated by XingGAN and XingGAN++, ensuring fairness by utilizing only the training data from the Market-1501 dataset. The results generated by both methods improved appearance consistency, yielding high-quality additional data. The augmented data is then integrated into the PAT \cite{ni2023part} training process, resulting in improvements in person re-identification performance in Table \ref{tab:data}.

\begin{table}[!t] \small
	\centering
	\caption{Performance results of person re-identification using PAT, with augmented data generated by XingGAN and XingGAN++ on the Market-1501 dataset.}
		\begin{tabular}{rcc} \toprule
			Data Source & R1 $\uparrow$ & mAP $\uparrow$ \\ \midrule
               PAT \cite{ni2023part} original data & 92.4 & 81.5 \\
               w/ augmented data by XingGAN & 92.8 & 82.0\\
               w/ augmented data by XingGAN++ & \textbf{93.5} & \textbf{82.4}\\
			\bottomrule	
	\end{tabular}
	\label{tab:data}
	\vspace{-0.4cm}
\end{table}

\begin{table}[!t] \small
	\centering
	\caption{User study of person image generation (\%). R2G means the percentage of real images rated as generated w.r.t. all real images. G2R means the percentage of generated images rated as real w.r.t. all generated images. The results of other methods are reported from their papers.}
		\begin{tabular}{rccccccc} \toprule
			\multirow{2}{*}{Method}  & \multicolumn{2}{c}{Market-1501} & \multicolumn{2}{c}{DeepFashion} \\ \cmidrule(lr){2-3} \cmidrule(lr){4-5} 
			& R2G $\uparrow$ & G2R $\uparrow$ & R2G $\uparrow$ & G2R $\uparrow$ \\ \hline	
			PG2~\cite{ma2017pose}                              & 11.2  & 5.5    & 9.2   & 14.9 \\
			PoseGAN~\cite{siarohin2018deformable}    & 22.67 & 50.24  & 12.42 & 24.61 \\ 
			C2GAN~\cite{tang2019cycle}                      & 23.20 & 46.70  & -     & -     \\
			PATN~\cite{zhu2019progressive}   & 32.23 & 63.47  & 19.14 & 31.78 \\  
                NTED \cite{ren2022neural} & 36.81 & 67.25 & 23.80 & 36.24 \\
			XingGAN \cite{tang2020xinggan} & 35.28 & 65.16 & 21.61 & 33.75 \\	
                XingGAN++ (Ours) & \textbf{38.98} & \textbf{69.43} & \textbf{25.88} & \textbf{38.76} \\
			\bottomrule	
	\end{tabular}
	\label{tab:pose_ruser}
	\vspace{-0.4cm}
\end{table}

\noindent \textbf{Qualitative Comparisons.}
The results compared with PG2~\cite{ma2017pose}, VUNet~\cite{esser2018variational},  PoseGAN~\cite{siarohin2018deformable}, XingGAN  \cite{tang2020xinggan}, and PIDM \cite{bhunia2023person} are shown in Figure~\ref{fig:market_result} (a) and~\ref{fig:fashion_result} (a), respectively. 
We can see that the proposed XingGAN++ achieves much better results than PG2, VUNet, PoseGAN, and XingGAN on both datasets, especially in terms of appearance details and the integrity of generated persons.
Moreover, to evaluate the effectiveness of XingGAN++, we compare it with five stronger approaches, i.e., PATN \cite{zhu2019progressive}, PoseStylizer~\cite{huang2020generating}, SPGNet \cite{lv2021learning}, NTED \cite{ren2022neural}, XingGAN  \cite{tang2020xinggan}, and PIDM \cite{bhunia2023person}.
The results are shown in Figure~\ref{fig:market_result} (b) and~\ref{fig:fashion_result} (b), respectively.
We can see that XingGAN++ again generates much better person images with fewer visual artifacts than PATN, PoseStylizer, SPGNet, NTED, and XingGAN.
For instance, PATN always generates a lot of visual artifacts in the background, as shown in Figure~\ref{fig:fashion_result} (b).

\noindent \textbf{User Study.}
Following the evaluation protocol of \cite{ma2017pose,siarohin2018deformable,zhu2019progressive}, we recruited 30 volunteers to conduct a user study.
Participants were shown a sequence of images and asked to judge each image in a second.
Specifically, we randomly selected 55 real and 55 fake images (generated by our model) and shuffled them.
The first ten were used for practice, and the remaining 100 images were used for evaluation.
The results compared with PG2~\cite{ma2017pose}, PoseGAN~\cite{siarohin2018deformable}, PATN~\cite{zhu2019progressive}, C2GAN \cite{tang2019cycle}, NTED \cite{ren2022neural}, and XingGAN \cite{tang2020xinggan} are shown in Table \ref{tab:pose_ruser}.
We observe that the proposed XingGAN++ achieves the best results on all measurements compared with the leading methods, further validating that the images generated by our models are sharper and more photorealistic.

\begin{table}[!t] \small
	\centering
	\caption{Training and inference time.}
		\begin{tabular}{rcc} \toprule
			Method & Training Time $\downarrow$ & Inference Time $\downarrow$ \\ \midrule
                PIDM \cite{bhunia2023person} & 104 days & 16.975 s \\
                X-MDPT-L \cite{pham2024cross} & 15 days & 3.124 s \\
                XingGAN++ (Ours) & \textbf{1.5 days} & \textbf{0.941 s} \\
			\bottomrule	
	\end{tabular}
	\label{tab:time}
	\vspace{-0.3cm}
\end{table}

\noindent \textbf{Training and Inference Time.} 
We also provide training and inference times in Table \ref{tab:time}. Our GAN-based method is much more efficient in training and inference time compared to both diffusion-based methods when using a single A100 GPU. Specifically, the proposed method is 69.33$\times$ faster than PIDM during training and 18.04$\times$ faster during inference. Moreover, the proposed method is also 10$\times$ faster than X-MDPT-L during training and 3.32$\times$ faster during inference.

\noindent \textbf{XingGAN vs. XingGAN++.}
We also compare XingGAN and XingGAN++ on both Market-1501 and DeepFashion datasets. 
The results are shown in Tables \ref{tab:pose_reuslts} and \ref{tab:pose_ruser}.
We see that XingGAN++ achieves much better results than XingGAN in all metrics, indicating that the proposed enhanced multi-scale cross-attention blocks learn the correlations between person poses at different scales, thus improving the generation performance.
From the visualization results in Figures~\ref{fig:market_result} and \ref{fig:fashion_result}, we can see that XingGAN++ generates more photorealistic images with fewer visual artifacts than XingGAN on both datasets.

\subsection{Ablation Study}
\noindent \textbf{Baselines.}
We conduct extensive ablation studies on Market-1501~\cite{zheng2015scalable} to evaluate different components of the proposed method.
The proposed method has six baselines (i.e., B1, B2, B3, B4, B5, B6), as shown in Table~\ref{tab:abla}: (1) `SA' means only the proposed shape-guided appearance-based generation branch is used. (2) `AS' means only the proposed appearance-guided shape-based generation branch is adopted. (3) `SA+AS' combines both branches to produce the final person images. (4) `SA+AS+CAF' employs the proposed co-attention fusion (CAF) module.
(5) `SA+AS+DCCAF' uses the proposed densely connected co-attention fusion (DCCAF) module instead of the co-attention fusion module.
(6) `EMSA+EMAS+DCCAF' is our full model and employs the proposed enhanced multi-scale cross-attention blocks.

\begin{table}[!t] \small
	\centering
	\caption{Quantitative comparison of different variants of the proposed method on the Market-1501 dataset.}
			\resizebox{1\linewidth}{!}{%
		\begin{tabular}{clcccc} \toprule
			\# & Baseline & IS $\uparrow$ & Mask-IS $\uparrow$ & SSIM $\uparrow$ & Mask-SSIM $\uparrow$    \\ \hline	
			B1 & SA & \textbf{3.849} & 3.645 & 0.239 & 0.768 \\ 
			B2 & AS & 3.796 & 3.810 & 0.286 & 0.798 \\
			B3 & SA + AS & 3.558 & 3.807 & 0.310 & 0.807 \\
			B4 & SA + AS + CAF & 3.506 & 3.872 & 0.313 & 
                0.816 \\
                B5 & SA + AS + DCCAF  & 3.522 & 3.879 & 0.317 & 0.819 \\
                B6 & EMSA + EMAS + DCCAF & 3.645 & \textbf{3.903} & \textbf{0.333} & \textbf{0.828}  \\ 
			\bottomrule	
	\end{tabular}}
	\label{tab:abla}
	\vspace{-0.4cm}
\end{table}

\begin{table}[!t] \small
	\centering
	\caption{Quantitative comparison of different variants of the proposed enhanced multi-scale cross-attention on the Market-1501 dataset.}
			\resizebox{1\linewidth}{!}{%
		\begin{tabular}{clcccc} \toprule
			Baseline & IS $\uparrow$ & Mask-IS $\uparrow$ & SSIM $\uparrow$ & Mask-SSIM $\uparrow$    \\ \hline	
			
                SA + AS + DCCAF  & 3.522 & 3.879 & 0.317 & 0.819 \\
                HSA + HAS + DCCAF & 3.524 & 3.881 & 0.318 & 0.820  \\ 
                MSA + MAS + DCCAF & 3.578 & 3.889 & 0.328 & 0.824  \\ 
                EMSA + EMAS + DCCAF & \textbf{3.645} & \textbf{3.903} & \textbf{0.333} & \textbf{0.828}  \\ 
			\bottomrule	
	\end{tabular}}
	\label{tab:abla2}
	\vspace{-0.4cm}
\end{table}

\begin{figure}[!t] \small
	\centering
	\includegraphics[width=1\linewidth]{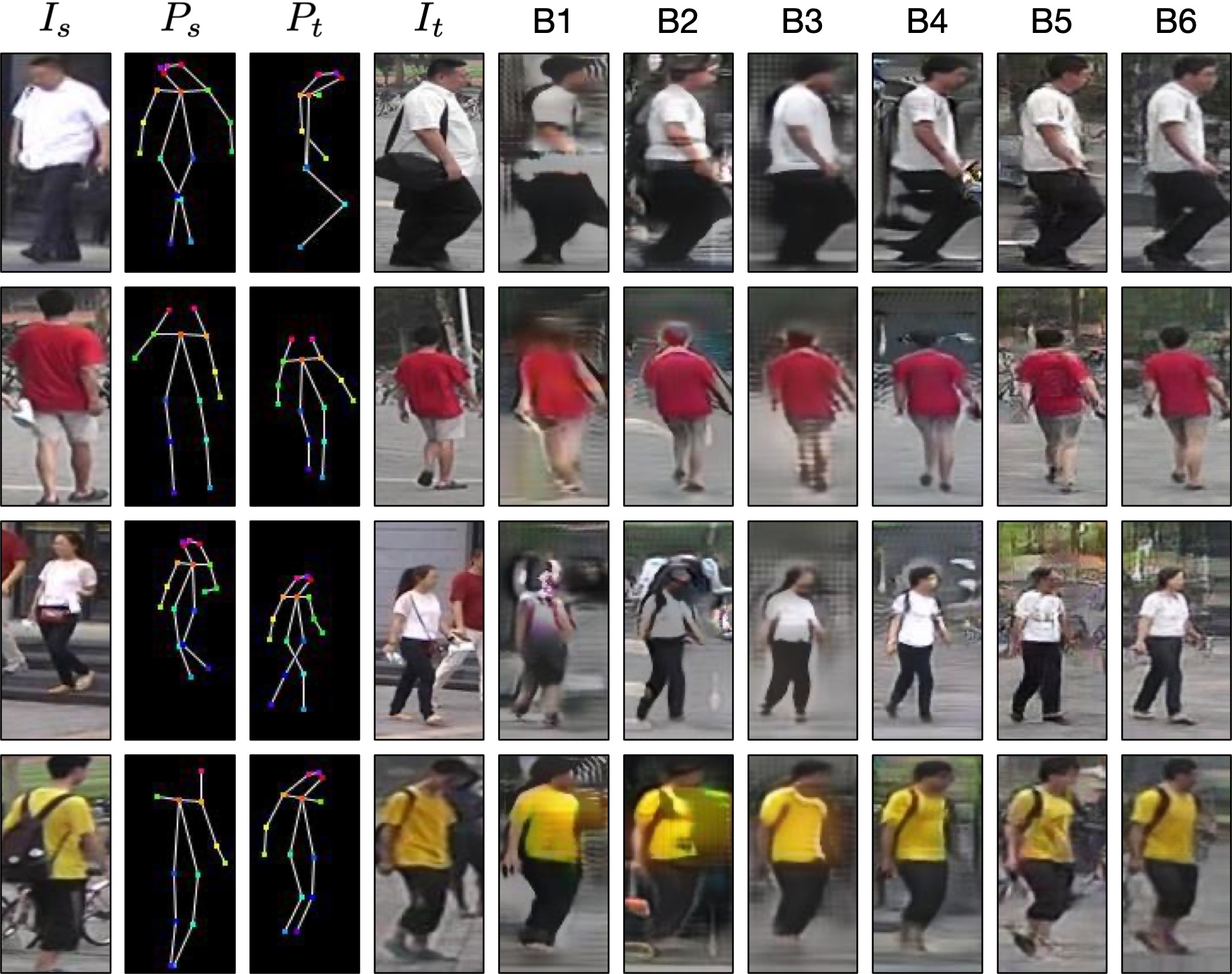}
	\caption{Ablation study of our method on Market-1501.}
	\label{fig:ablation}
	\vspace{-0.4cm}
\end{figure}

\noindent \textbf{Effect of Dual-Branch Generation.}
The results of the ablation study are shown in Table \ref{tab:abla}. 
As can be seen, the proposed SA branch achieves only 0.239 and 0.768 in SSIM and Mask-SSIM, respectively.
When we only use the proposed AS branch, the SSIM and Mask-SSIM values are improved to 0.286 and 0.798, respectively.
Thus, we conclude that the AS branch is more effective than the SA branch for generating photorealistic person images.
The AS branch takes the person poses as input and aims to learn the person's appearance representations, while the SA branch takes the person image as input and aims to learn person shape representations.
Learning the appearance representations is much easier than learning the shape representations since shape deformations occur between the input and desired person image, leading the AS branch to achieve better results than the SA branch.

Next, when adopting a combination of the proposed SA and AS branches, the performance in terms of SSIM and Mask-SSIM further improves.
Meanwhile, the results in terms of IS and Mask-IS do not decline too much.
Figure~\ref{fig:ablation} shows some qualitative examples of the ablation study.
We observe that the visualization results of `SA', `AS', and `SA+AS' are consistent with the quantitative results.
Therefore, both quantitative and qualitative results confirm the effectiveness of the proposed dual-branch generation strategy.

\begin{figure}[t] \small
	\centering
	\includegraphics[width=0.6\linewidth]{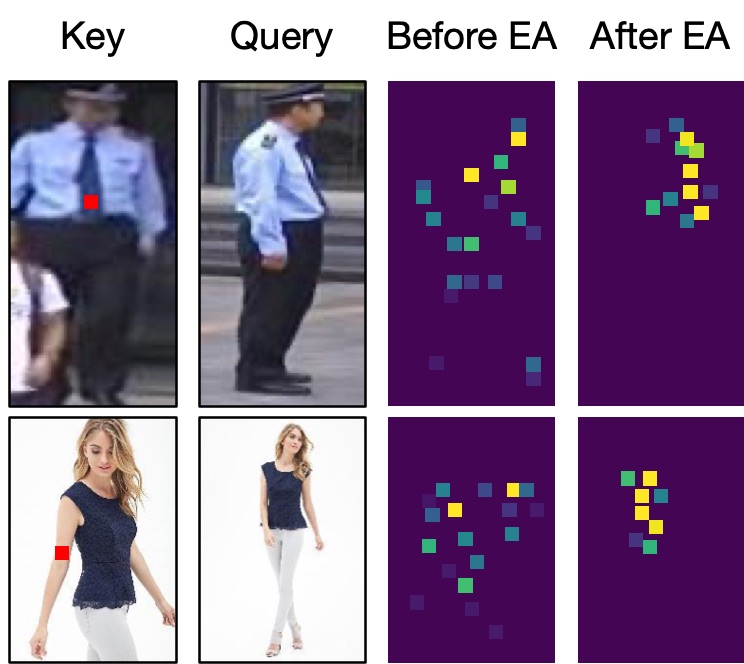}
	\caption{
Visualization of the impact of the introduced EA module. We display two correlation vectors, illustrating their states before and after undergoing refinement via the EA module. Specifically, we focus on the correlation vectors of keys associated with the target object regions presented in the third column. Notably, the EA module proves effective in suppressing erroneous correlations while enhancing the relevant ones.}
	\label{fig:ea}
	\vspace{-0.4cm}
\end{figure}

\begin{figure*}[!t] \small
	\centering
	\includegraphics[width=1\linewidth]{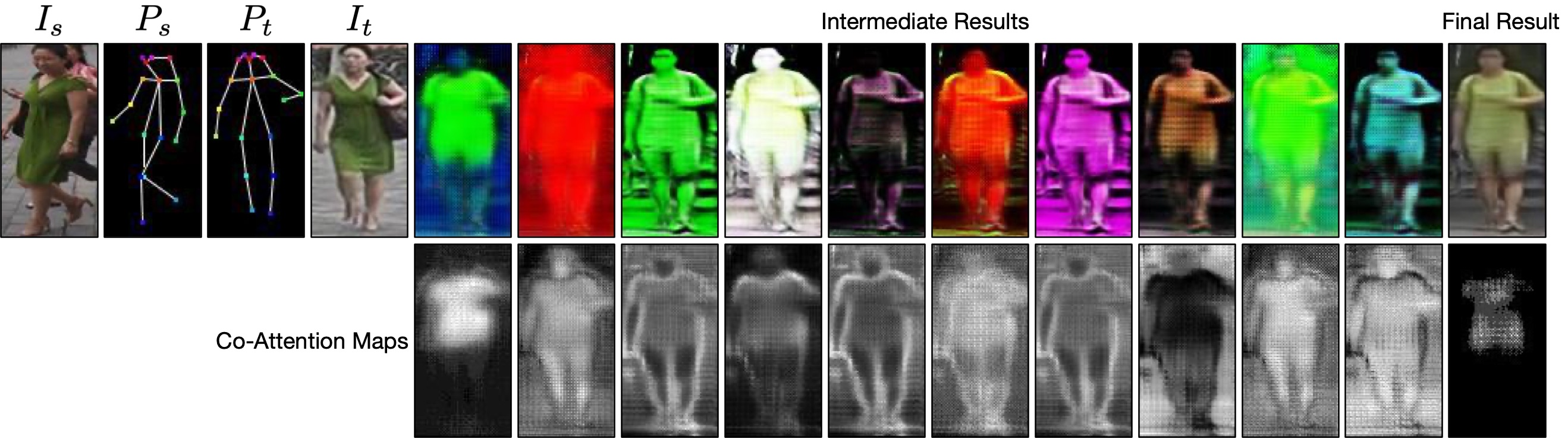}
	\caption{Visualization of intermediate results and co-attention maps generated by the proposed method on Market-1501. We show ten random intermediate results, the ten corresponding co-attention maps, and the input attention map. Attention maps are normalized for better visualization.}
	\label{fig:attention_map}
	\vspace{-0.2cm}
\end{figure*}

\begin{table*}[!t] \small
	\centering
	\caption{Quantitative comparison and ablation study of the proposed Xing and Xing++ generators on Market-1501.}
		\begin{tabular}{lcccc} \toprule
			Method                      & IS $\uparrow$    & Mask-IS $\uparrow$ & SSIM $\uparrow$  & Mask-SSIM $\uparrow$ \\ \hline
			Xing Generator (1 blocks) & 3.378 & 3.713   & 0.310  & 0.812 \\
			Xing Generator (3 blocks)  & 3.241 & 3.866   & 0.316  & 0.813 \\
			Xing Generator (5 blocks)   & 3.292 & 3.860  & 0.313  & 0.812 \\ 
			Xing Generator (7 blocks)   & 3.293 & 3.871     & 0.310  & 0.810 \\
			Xing Generator (9 blocks) & 3.506 & 3.872 & 0.313  & 0.816  \\
			Xing Generator (11 blocks)  & 3.428 & 3.712     & 0.286  & 0.793 \\
			Xing Generator (13 blocks)  & \textbf{3.708}    & 3.679     & 0.257  & 0.774 \\ \hline
                Xing++ Generator (1 blocks) & 3.452 & 3.787 & 0.318 & 0.817 \\
			Xing++ Generator (3 blocks)  & 3.503 & 3.872 & 0.326 & 0.822 \\
			Xing++ Generator (5 blocks)   & 3.645 & \textbf{3.903} &\textbf{0.333} & \textbf{0.828}  \\ 
			Xing++ Generator (7 blocks)   &  3.642 & 3.882 & 0.325 & 0.823 \\
			Xing++ Generator (9 blocks) & 3.691 & 3.880 & 0.320 & 0.819 \\
			Xing++ Generator (11 blocks)  & 3.587 &3.756 & 0.315 & 0.814 \\
			Xing++ Generator (13 blocks)  & 3.697 & 3.735 & 0.309 & 0.808 \\ \hline
                
			ResNet Generator (5 blocks) & 3.236 & 3.807     & 0.297  & 0.802 \\
			ResNet Generator (9 blocks) & 3.077 & 3.862     & 0.301 &  0.802 \\
			ResNet Generator (13 blocks)& 3.134 &  3.731    & 0.300 & 0.797 \\ \hline
			PATN Generator (5 blocks)   & 3.273 & 3.870     & 0.309 &  0.809 \\
			PATN Generator (9 blocks)   &  3.323 & 3.773    & 0.311 & 0.811 \\
			PATN Generator (13 blocks)  & 3.274 & 3.797     & 0.314 & 0.808 \\  \hline
                Self-Attention Generator (5 blocks)   &  3.265 & 3.796 & 0.304 & 0.806 \\
			Self-Attention Generator (9 blocks)   & 3.465 & 3.786 & 0.308 & 0.809  \\
			Self-Attention Generator (13 blocks)  & 3.676 & 3.596 & 0.248 & 0.766 \\ 
			\bottomrule	
	\end{tabular}
	\label{tab:ablation_blocks}
	\vspace{-0.4cm}
\end{table*}

\noindent \textbf{Effect of Enhanced Multi-Scale Cross-Attention.}
We see that B6 is much better than B5, clearly proving the effectiveness of the proposed enhanced multi-scale cross-attention blocks.
Moreover, we provide more detailed comparison results for the proposed enhanced multi-scale cross-attention in Table \ref{tab:abla2}.
`SA+AS+DCCAF' employs the SA and AS blocks.
`MSA+MAS+DCCAF' uses the proposed multi-scale cross-attention blocks.
Compared with `MSA+MAS+DCCAF', `HSA+HAS+DCCAF' uses multi-head attention instead of the proposed multi-scale attention.
`EMSA+EMAS+DCCAF' uses the enhanced multi-scale cross-attention blocks.
We observe that the results of `HSA+HAS+DCCAF' and `SA+AS+DCCAF' are similar, indicating that multi-head attention cannot continue to improve the results.
Moreover, we see that `MSA+MAS+DCCAF' is better than `SA+AS+DCCAF' and `HSA+HAS+DCCAF' on all metrics, illustrating the effectiveness of our proposed multi-scale cross-attention.
After adding the proposed enhanced attention (EA) module to `EMSA+EMAS+DCCAF', all metrics further improved compared with `MSA+MAS+DCCAF', thus verifying the effectiveness of our method EA module.
We also provide the visualization results of our EA module in Figure \ref{fig:ea}. We see that the EA module boosts relevant correlations and diminishes erroneous ones by fostering agreement among all correlation vectors.

\noindent \textbf{Effect of Densely Connected Co-Attention Fusion.}
`SA+AS+CAF' outperforms the `SA+AS' baseline with gains of around 0.065, 0.003, and 0.009 in Mask-IS, SSIM, and Mask-SSIM, respectively. 
This indicates that the proposed co-attention fusion model does indeed learn more correlations between the appearance and shape representations for generating the target person images, validating our design motivation.
Also, the proposed CAF module obviously improves the quality of the visualization results, as shown in Figure~\ref{fig:ablation}.

Moreover, B5 achieves better results than B4, which indicates that using the proposed densely connected co-attention fusion blocks to fuse the appearance and shape features at different stages will improve the generation performance.
For example, in the third row of Figure~\ref{fig:ablation}, B5 tries to generate a real background, while B4 can only generate a blurred background.

Lastly, we show the learned co-attention maps and the generated intermediate results.
Specifically, we show ten randomly chosen intermediate results, their ten corresponding co-attention maps, and the input attention map.
These co-attention maps are complementary, which can be qualitatively verified through the visualization results in Figure~\ref{fig:attention_map}. 
The model learns different activated content between the generated intermediate results and the input image for generating the final person images, revealing the effectiveness of the proposed co-attention fusion module.

\noindent \textbf{Effect of The Xing Generator.}
The proposed Xing generator has two important network designs. 
One is the carefully crafted Xing block, which consists of two sub-blocks, i.e., the SA block and the AS block. 
The Xing blocks jointly model both the shape and appearance representations through a cross strategy, mutually benefitting each other.
The other important component is the cascaded network design, which deals with the complex and deformable translation problem progressively.
Thus, we conduct two further experiments; one to demonstrate the advantage of the progressive generation strategy by varying the number of Xing blocks, and the other to explore the advantage of the Xing block by replacing it with the residual block \cite{johnson2016perceptual}, PATB \cite{zhu2019progressive}, and self-attention \cite{wang2018non}. This results in three generators, named the ResNet generator, PATN generator, and self-attention generator, respectively, as shown in Table \ref{tab:ablation_blocks}.

We observe that the proposed Xing generator with nine blocks works the best. 
However, further increasing the number of blocks reduces the generation performance.
This is due to the proposed Xing block, as only a few blocks are needed to capture the useful appearance and shape representations and the connection between them.
Thus, we adopt nine Xing blocks as default in our experiments for both datasets.
Moreover, the proposed Xing generator with only five Xing blocks outperforms the ResNet and PATN generators with 13 blocks in most metrics. This demonstrates that our Xing generator has a good appearance and shape modeling capabilities with few blocks.
We also observe that the proposed Xing generator achieves much better than the self-attention generator, which indicates single modality attention is not sufficient, and long-range cross-attention is needed for this challenging generation task.

\noindent \textbf{Xing vs. Xing++ Generator.}
We also compare the proposed Xing++ generator with the Xing generator in Table \ref{tab:ablation_blocks}.
The Xing++ generator with five blocks achieves the best results. However, further increasing the number of blocks reduces the generation performance.
Thus, we adopt five Xing++ blocks as default in XingGAN++ for both datasets. Moreover, we see that the Xing++ generator with only five blocks outperforms the Xing generator with nine blocks in all the metrics, which further demonstrates that the Xing++ generator has a good appearance and shape modeling capabilities with very few blocks than the Xing generator.

%% file: 5conclusions.tex
\section{Conclusion}
We propose two novel cross-attention-based models, i.e., XingGAN and XingGAN++, for the challenging person image generation task.
Specifically, XingGAN uses cascaded guidance with two different generation branches and learns a deformable translation mapping from both person shape and appearance features.
Moreover, we propose two novel blocks in XingGAN to effectively update the person's shape and appearance features through a cross operation.
Meanwhile, XingGAN++ utilizes the proposed multi-scale cross-attention modules to capture the long-range correlations between person poses at different scales.
We also propose a new enhanced attention (EA) module to improve accurate correlations and reduce misleading ones by establishing consensus among all correlation features.
Lastly, we introduce a novel densely connected co-attention fusion module to fuse appearance and shape features at different stages effectively.
Extensive experiments based on human assessment and automatic evaluation metrics show that XingGAN and XingGAN++ achieve new state-of-the-art results on two public datasets.
We believe the proposed modules can be easily extended to address other generation (e.g., virtual try-on) and even multi-modal fusion tasks.

\noindent \textbf{Acknowledgments.}
This work is partially supported by the Fundamental Research Funds for the Central Universities, Peking University. Nicu Sebe acknowledges funding by the European Union’s Horizon Europe research and innovation program under grant agreement No. 101120237 (ELIAS) as well as the support of the PNRR project FAIR - Future AI Research (PE00000013), under the NRRP MUR program funded by the NextGenerationEU.